\documentclass[10pt,twocolumn,letterpaper]{article}

\usepackage{cvpr}
\usepackage{times}
\usepackage{epsfig}
\usepackage{graphicx}
\usepackage{amsmath}
\usepackage{amssymb}
\usepackage{multirow}
\usepackage{boldline}

\usepackage[pagebackref=true,breaklinks=true,letterpaper=true,colorlinks,bookmarks=false]{hyperref}

\newcolumntype{x}[1]{>{\centering\arraybackslash}p{#1pt}}

\newlength\savewidth\newcommand\shline{\noalign{\global\savewidth\arrayrulewidth
  \global\arrayrulewidth 1pt}\hline\noalign{\global\arrayrulewidth\savewidth}}
\newcommand{\tablestyle}[2]{\setlength{\tabcolsep}{#1}\renewcommand{\arraystretch}{#2}\centering\footnotesize}
\makeatletter\renewcommand\paragraph{\@startsection{paragraph}{4}{\z@}
  {.5em \@plus1ex \@minus.2ex}{-.5em}{\normalfont\normalsize\bfseries}}\makeatother

\cvprfinalcopy 

\def\cvprPaperID{7045} 

\ifcvprfinal\fi
\begin{document}

\title{Rethinking Differentiable Search for Mixed-Precision Neural Networks}

\author{Zhaowei Cai\\
UC San Diego\\
{\tt\small zwcai@ucsd.edu}
\and
Nuno Vasconcelos\\
UC San Diego\\
{\tt\small nuno@ucsd.edu}
}

\maketitle

\begin{abstract}
Low-precision networks, with weights and activations quantized to low bit-width, are widely used to accelerate inference on edge devices. However, current solutions are uniform, using identical bit-width for all filters. This fails to account for the different sensitivities of different filters and is suboptimal. Mixed-precision networks address this problem, by tuning the bit-width to individual filter requirements. In this work, the problem of optimal mixed-precision network search (MPS) is considered. To circumvent its difficulties of discrete search space and combinatorial optimization, a new differentiable search architecture is proposed, with several novel contributions to advance the efficiency by leveraging the unique properties of the MPS problem. The resulting Efficient differentiable MIxed-Precision network Search (EdMIPS) method is effective at finding the optimal bit allocation for multiple popular networks, and can search a large model, e.g. Inception-V3, directly on ImageNet without proxy task in a reasonable amount of time. The learned mixed-precision networks significantly outperform their uniform counterparts.
\end{abstract}

\section{Introduction}
\label{sec:intro}

Deep neural networks have state-of-the-art performance on computer vision tasks such as visual recognition \cite{DBLP:conf/nips/KrizhevskySH12,DBLP:journals/corr/SimonyanZ14a,DBLP:conf/cvpr/SzegedyLJSRAEVR15,DBLP:conf/cvpr/SzegedyVISW16,DBLP:journals/corr/HeZRS15}, object detection \cite{DBLP:conf/nips/RenHGS15,DBLP:conf/cvpr/LinDGHHB17,DBLP:conf/cvpr/CaiV18}, segmentation \cite{DBLP:conf/iccv/HeGDG17,DBLP:journals/pami/ChenPKMY18}, etc. However, their large computation and memory costs make them difficult to deploy on devices such as mobile phones, drones, autonomous robots, etc. Low-precision networks, which severely reduce computation and storage by quantizing network weights and activations to low-bit representations, promise a solution to this problem.

In the low-precision literature, all network weights and activations are usually quantized to the same bit-width \cite{DBLP:conf/nips/HubaraCSEB16,DBLP:conf/eccv/RastegariORF16,DBLP:journals/corr/ZhouNZWWZ16,DBLP:conf/cvpr/CaiHSV17,DBLP:conf/cvpr/ZhuangSTL018,DBLP:conf/eccv/ZhangYYH18}. The resulting {\it uniform\/} low-precision networks have been preferred mostly because they are well supported by existing hardware, e.g. CPUs, FPGAs, etc. However, uniform bit allocation does not account for the  individual properties of different filters, e.g. their location on the network, structure, parameter cardinality, etc. In result, it can lead to suboptimal performance for a given network size and complexity.  {\it Mixed}-precision networks \cite{DBLP:conf/icassp/AnwarHS15,DBLP:conf/icml/LinTA16,DBLP:conf/uai/LaceyTA18,yazdanbakhsh2018releq,wu2018mixed,wang2019haq,DBLP:conf/iccv/DongYGMK19}
address this limitation, enabling the optimization of bit-widths at the filter level.
They are also becoming practically relevant, with the introduction of hardware that supports mixed-precision representations. For example, the NVIDIA AMP\footnote{https://devblogs.nvidia.com/nvidia-automatic-mixed-precision-tensorflow} can
choose between different floating point representations during training.

\begin{figure}[!t]
\centering
\centerline{\epsfig{figure=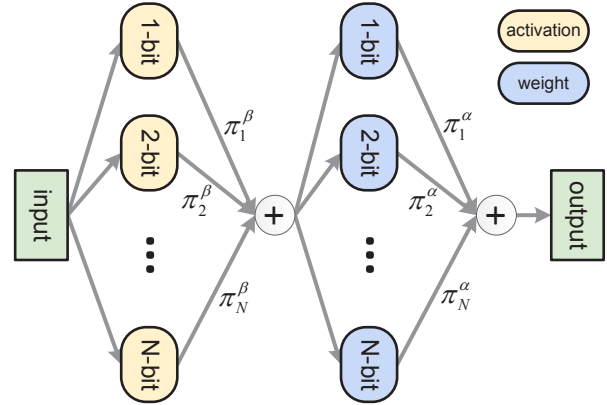,width=8cm,height=5.4cm}}
\caption{Differentiable architecture of the proposed mixed-precision network search module.}
\label{fig:mixprecision}
\end{figure}

Nevertheless, the problem of optimizing bit allocation for a mixed-precision network is very challenging. Since a network of $L$ layers and
$N$ candidate bit-widths can have $N^L$ different configurations, it is usually impossible to manually craft the optimal solution, and automatic bit allocation techniques are required. While this is well aligned with recent progresses in
automatic neural architecture search (NAS)
\cite{DBLP:journals/corr/ZophL16,DBLP:conf/cvpr/ZophVSL18,DBLP:conf/iclr/BrockLRW18,DBLP:conf/icml/BenderKZVL18,liu2018darts},
there are several important differences between generic NAS and mixed-precision network search (MPS). First, NAS relies extensively on a proxy task
to overcome the extremely high computational demands of searching for the optimal network architecture on a large dataset, such as ImageNet. It is common to search for a module block on a small dataset
(such as CIFAR-10) and stack copies of this module as the final architecture.
However, this type of proxy task is very ineffective for MPS, due to 1) the layer importance difference,
e.g. layers closer to input and output typically require higher bit-width;
and 2) the (likely large) difference between optimal bit allocations for CIFAR-10 and ImageNet. Second, higher bit-widths usually lead (up to overfitting)
to mixed-precision networks of higher accuracy. Hence, the sole
minimization of classification loss usually has a trivial solution: to always select the candidate of highest bit-width. Third, while general NAS requires candidate operators that are heterogeneous in structure, e.g. convolution, skip connection, pooling, etc., MPS only involves homogeneous operators and very
similar representations, e.g. convolutions of different bit-widths.
These unique properties of MPS suggest the need for search techniques different than those of
standard NAS.

In this work, we leverage the above properties to propose an efficient MPS framework based on several contributions. First, to enable search without proxies, the proposed framework is based on the differentiable search architecture of Figure \ref{fig:mixprecision}, motivated by the popular DARTS \cite{liu2018darts} approach to generic NAS. Second, to avoid the trivial selection of the highest bit-width, learning is constrained by a complexity budget. The constrained optimization is reformulated as a Lagrangian, which is optimized to achieve the optimal trade-off between accuracy and complexity. Third, to circumvent the expensive second-order bi-level optimization of DARTS, a much simpler and effective optimization is proposed, where both architecture and network parameters are updated in a single forward-backward pass.
Fourth, by exploiting the linearity of the convolution operator, the expensive parallel convolutions of Figure \ref{fig:mixprecision} are replaced by an efficient composite convolution, parameterized by the weighted sum of parallel weight tensors.
This ensures the training complexity remains constant, independently of the size of search space, enabling the training of large networks.

Together, these contributions enable an efficient MPS procedure with no need for proxy tasks. This is denoted as {\it Efficient differentiable MIxed-Precision network Search\/} (EdMIPS) and can, for example, search the optimal mixed-precision Inception-V3 \cite{DBLP:conf/cvpr/SzegedyVISW16} of 93 filters, on ImageNet, in 8 GPU days. Extensive evaluations of EdMIPS on multiple popular networks of various sizes, accuracies, properties, etc., including AlexNet, ResNet, GoogLeNet and Inception-V3, show that it outperforms uniform low-precision solutions by a large margin. Beyond demonstrating the effectiveness of EdMIPS, this vast set of results also establishes solid baselines for the growing area of mixed-precision networks. To facilitate future research, all code is released at \url{https://github.com/zhaoweicai/EdMIPS}.

\section{Related Work}

\noindent{\bf Uniform low-precision:} Low-precision networks have recently become popular to speed-up and reduce model size of deep networks \cite{DBLP:conf/nips/HubaraCSEB16,DBLP:conf/eccv/RastegariORF16,DBLP:journals/corr/ZhouNZWWZ16,DBLP:conf/cvpr/CaiHSV17,DBLP:conf/cvpr/ZhuangSTL018,DBLP:conf/eccv/ZhangYYH18,zhuang2019structured}. \cite{DBLP:conf/nips/HubaraCSEB16,DBLP:conf/eccv/RastegariORF16} pioneered the joint binarization of network weights and activations, using a  continuous approximation to overcome the non-differentiability of quantization. This, however, entailed a significant accuracy loss.
\cite{DBLP:journals/corr/ZhouNZWWZ16,DBLP:conf/cvpr/CaiHSV17} later achieved accuracies much closer to those of full-precision networks. The HWGQ-Net \cite{DBLP:conf/cvpr/CaiHSV17} approximated the ReLU by a half-wave Gaussian quantizer, and proposed a clipped ReLU function to avoid the gradient mismatch. LQ-Net \cite{DBLP:conf/eccv/ZhangYYH18} and PACT \cite{choi2018pact} tried to learn the optimal step size and clipping function online, respectively, achieving better
performance.
However, these are all uniform low-precision networks.

\noindent{\bf Bit allocation:} Optimal bit allocation has a long history in neural networks \cite{DBLP:conf/sips/HwangS14,DBLP:conf/icassp/AnwarHS15,DBLP:conf/icml/LinTA16}. \cite{DBLP:conf/sips/HwangS14,DBLP:conf/icassp/AnwarHS15} proposed the exhaustive checking of filter sensitivities on a per layer basis, following simulation-based word-length optimization methods from signal processing \cite{DBLP:journals/tsp/SungK95}.
\cite{DBLP:conf/icml/LinTA16}
proposed an analytical solution to fixed-point quantization that seeks the optimal bit-width allocations across network layers by optimizing signal-to-quantization-noise-ratio (SQNR). \cite{DBLP:conf/uai/LaceyTA18} frames precision allocation as the sequential allocation of bits of precision to the weights of one layer, until a bit budget is exhausted. These techniques precede NAS \cite{DBLP:journals/corr/ZophL16} and are less powerful, less applicable to practical network design, or both.

\noindent{\bf Neural architecture search:} NAS is a popular approach to automated search of neural network architectures \cite{DBLP:journals/corr/ZophL16,DBLP:conf/cvpr/ZophVSL18,DBLP:conf/iclr/BrockLRW18,DBLP:conf/icml/BenderKZVL18,liu2018darts}. However, the large and discrete nature of the search space make NAS very expensive. \cite{DBLP:journals/corr/ZophL16} proposed a reinforcement learning (RL) technique that requires 1,000 GPU days to search for an architecture on CIFAR-10. Subsequent works have tried to reduce these extraordinary levels of computation. \cite{DBLP:conf/iclr/BrockLRW18,DBLP:conf/icml/BenderKZVL18,guo2019single} start by learning a supernet, containing all possible module choices, and find the best subnet within this supernet. Differentiable architecture search (DARTS) \cite{liu2018darts} relaxes the discrete search space into a continuous one, enabling the optimization by gradient descent.

\noindent{\bf Mixed Precision:} Recently, \cite{yazdanbakhsh2018releq,wu2018mixed,wang2019haq} formulated
MPS as an instance of NAS. Some of these
techniques \cite{yazdanbakhsh2018releq,wang2019haq} are based on RL and thus
not very efficient.
\cite{wu2018mixed} relies on DARTS but requires a proxy task, due to the linear memory/computation on the cardinality of candidate pool, and samples dozens of architectures during search. Compared to these
approaches, EdMIPS is simpler, more efficient, effective, and applicable
to a much broader set of networks.

\section{Low-Precision Neural Network}

In this section, we briefly review some preliminary concepts on low-precision
neural network.

\subsection{Deep Neural Network}

Deep networks implement many filtering operators,
\begin{equation}
\label{equ:conv}
y=f(a(x)) = \mathbf{W}\ast{a}(x),
\end{equation}
where filter $f$ is parameterized by weight tensor $\textbf{W}$
(which can be expressed in full or low-precision), $x$ is the filter input, $a$ a non-linear
activation function (e.g. ReLU in full- or HWGQ \cite{DBLP:conf/cvpr/CaiHSV17} in low-precision
networks), $y$ the filter output and $\ast$ the convolution or matrix-vector multiplication operator. The $L$ network filters $F=\{f_1,f_2,\cdots,f_L\}$ are
learned by minimizing the classification risk ${\cal R}_{E}[F]$ defined by the cross-entropy
loss function
on a training set. The overall complexity of a neural network is mostly dictated by the complexity of (\ref{equ:conv}).
By quantizing both weights and activations to low bit-width, the expensive float-point operations of (\ref{equ:conv}) can be replaced by efficient bit operations (e.g. XNOR and bit-count), substantially reducing model size and computation \cite{DBLP:conf/eccv/RastegariORF16,DBLP:conf/cvpr/CaiHSV17}.

\begin{figure}[!t]
\centering
\centerline{\epsfig{figure=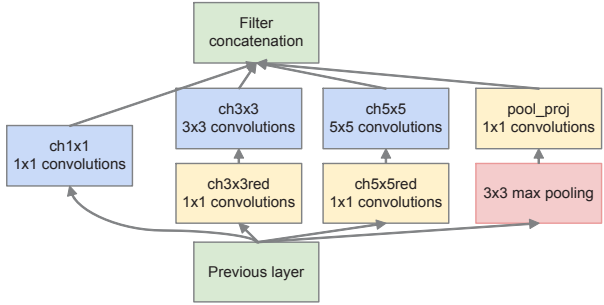,width=8cm,height=4.08cm}}
\caption{The Inception module in GoogLeNet.}
\label{fig:inception}
\end{figure}

\subsection{HWGQ Network}

A mixed-precision network can have arbitrary weight and activation bit-widths.
Many quantization techniques have been proposed \cite{DBLP:conf/nips/HubaraCSEB16,DBLP:conf/eccv/RastegariORF16,DBLP:journals/corr/ZhouNZWWZ16,DBLP:conf/cvpr/CaiHSV17,DBLP:conf/cvpr/ZhuangSTL018,DBLP:conf/eccv/ZhangYYH18}. In this work, we start from the HWGQ-Net \cite{DBLP:conf/cvpr/CaiHSV17}, one of the state-of-the-art low-precision networks in the literature.
Although the HWGQ-Net only uses binary weights, its activation quantization technique can be used to produce weights of higher precision. This starts by pre-computing the optimal quantizer
\begin{equation}
  Q(x)=q_i, \quad if \quad x\in(t_i,t_{i+1}]
  \label{eq:Q}
\end{equation}
for a Gaussian distribution of zero mean and unit variance,
using Lloyd's
algorithm \cite{DBLP:journals/tit/Lloyd82,DBLP:journals/tit/Max60}.
Since the network weight distributions are always close to zero-mean
Gaussians of different variance $\sigma^2$, the optimal quantization
parameters can then be easily obtained by rescaling the pre-computed quantization
parameters \cite{sun2008image}, i.e. using the quantizer
\begin{equation}
Q(x)=\sigma q_i, \quad if \quad x\in(\sigma t_i,\sigma t_{i+1}].
\end{equation}
To be hardware-friendly, all quantizers are uniform.

\begin{figure}[!t]
\centering
\centerline{\epsfig{figure=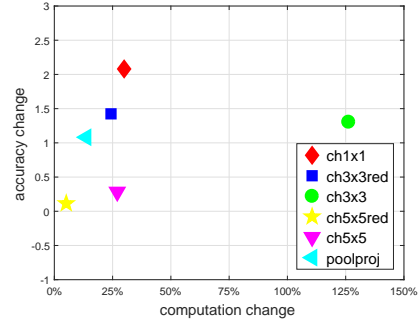,width=6cm,height=4.5cm}}
\caption{The filter sensitivity in the Inception module.}
\label{fig:sensitivity}
\end{figure}

\subsection{Filter Sensitivity}

The allocation of the same bit-width to all network layers, known as uniform
bit allocation, can be very suboptimal, since different filters have
different bit-width sensitivities. Figure \ref{fig:sensitivity}
illustrates this point for the popular Inception module of GoogLeNet \cite{DBLP:conf/cvpr/SzegedyLJSRAEVR15} whose architecture is shown in Figure \ref{fig:inception}. This has four parallel branches and six learnable filters in total.
To check their sensitivities, we started by training a uniform 2-bit
GoogLeNet baseline on ImageNet. We then trained another model, changing
a single filter of the Inception module to 4-bit, throughout
the network. This experiment was repeated for all six filters. The
changes in accuracy and computation of the whole network,
with respect to the baseline, reflect the differences in bit-width
sensitivity of the six filters. Figure \ref{fig:sensitivity}
shows that ``ch1x1'' is the most sensitive filter, as
the bit-width increase improves network accuracy by more than 2\%
with only a 25\% computation increase. ``ch3x3red'' and ``ch5x5'' have computation  similar to ``ch1x1'' but are less sensitive, especially ``ch5x5''. On the other hand, the computation
of ``ch3x3'' skyrockets to 125\% for a much smaller gain of 1.3\% in accuracy,
which is close to that of the inexpensive ``ch3x3red'' and ``pool\_proj''. Finally, ``ch5x5red'' has minor changes in accuracy and computation.
These observations suggest that mixed-precision networks could
substantially outperform uniform networks.

\section{Mixed-Precision Network}

In a mixed-precision network, bit-width varies filter by filter. The
problem is to search a candidate pool $\cal{B}$ of bit-widths
for the optimal bit-width for each filter in the network. This can be implemented
by reformulating (\ref{equ:conv}) as
\begin{align}
\label{equ:hard mix}
  &y=\sum_{i=1}^{n_f}o^\alpha_i f_i\left(\sum_{j=1}^{n_a}o^\beta_j
    a_j(x)\right), \\\nonumber
  \mbox{s.t. } \sum &o_i^\alpha=1,\sum o_j^\beta=1,o^\alpha,o^\beta\in\{0,1\},
\end{align}
where $n_f$ and $n_a$ are the cardinalities of bit-width pool $\cal{B}^\alpha$ for weights and $\cal{B}^\beta$ for activations. The goal is to find the optimal bit-width configuration $\{o_\alpha^*,o_\beta^*\}$ for the whole network. Since the search space is discrete and large, it is usually infeasible to hand-craft the optimal solution.

\subsection{Complexity-Aware Learning}

In general, networks of higher bit-width have higher accuracy. Hence, the simple minimization of classification loss has the trivial solution of always selecting the highest possible bit-width. To avoid this,
we resort to complexity-aware learning, seeking the best trade-off between classification accuracy and complexity. This is a constrained
optimization problem, where classification risk ${\cal R}_{E}[F]$ is minimized under a bound
on complexity risk $R_C[F]$,
\begin{equation}
\label{equ:constrained optimization}
F^* = \arg\min_F R_E[F] \quad  \mbox{s.t.} \quad R_C[F] < \gamma,
\end{equation}
which can be solved by minimizing the Lagrangian
\begin{equation}
\label{equ:lagrangian}
\mathcal{L}[F]=\mathcal{R}_{E}[F]+\eta\mathcal{R}_{C}[F],
\end{equation}
where $\eta$ is a Lagrange multiplier that only depends on $\gamma$. Under this constrained formulation, the optimal bit allocation is no longer trivial.
The complexity is user-defined, and could address computation, memory, model size, energy,
running speed, or other aspects.
It has the form
\begin{equation}
\mathcal{R}_{C}[F]=\sum_{f\in{F}}c(f).
\label{eq:comprisk}
\end{equation}
where $c(f)$ is the cost of filter $f$.

\subsection{Model Complexity}

A popular practice is to characterize complexity by the number of floating-point operations (FLOPs) of filter $f$,
\begin{equation}
c(f)=|f|w_x h_x/s^2,
\label{equ:complexity}
\end{equation}
where $|\cdot|$ denotes cardinality, $w_x$ and $h_x$ are the spatial width and height of the filter input $x$, and $s$ the filter stride. In low-precision networks, where filter $f$ and activation function $a$ have low bit-width, this cost can be expressed in bit operations (BitOps),
\begin{equation}
c(f)=b_f b_a|f|w_x h_x/s^2,
\label{equ:bit complexity}
\end{equation}
where $b_f$ and $b_a$ are the bit-widths of weights and activations, respectively.
Since only relative complexities matter for the search, the overall network complexity is normalized by the value of (\ref{equ:complexity}) for the first layer to search.

\subsection{Relaxed Mixed-Precision Network}
\label{subsec:relexation}

The binary nature of the search space of $\{o_\alpha,o_\beta\}$ makes the minimization of (\ref{equ:lagrangian}) a complex combinatorial problem. As suggested by \cite{liu2018darts}, a much simpler optimization is possible by relaxing the binary search space into a continuous one, through the reformulation of (\ref{equ:hard mix}) as
\begin{align}
\label{equ:soft mix}
&y=\sum_{i=1}^{n_f}\pi^\alpha_i f_i\left(\sum_{j=1}^{n_a}\pi^\beta_j a_j(x)\right), \\\nonumber
\mbox{s.t.} \sum &\pi_i^\alpha=1,\sum \pi_j^\beta=1,\pi^\alpha,\pi^\beta\in[0,1].
\end{align}
The constraints $\pi^\alpha,\pi^\beta\in[0,1]$ can then be eliminated by introducing a set of real configuration parameters $\{\alpha, \beta\}$ and defining
\begin{equation}
\pi^\alpha_i=\frac{\exp(\alpha_i)}{\sum_k \exp(\alpha_k)},\quad \pi^\beta_j=\frac{\exp(\beta_j)}{\sum_k \exp(\beta_k)}.
\label{equ:pi}
\end{equation}
This leads to the architecture of Figure~\ref{fig:mixprecision}.
The complexity measure of (\ref{equ:bit complexity}) is finally defined as
\begin{equation}
c(f)=E[b_f]E[b_a]|f|w_x h_x/s^2,
\label{equ:soft complexity}
\end{equation}
where
\begin{equation}
E[b_f]=\sum_{i=1}^{n_f}\pi^\alpha_i b_{f_i},\quad E[b_a]=\sum_{j=1}^{n_a}\pi^\beta_j b_{a_j}
\label{equ:bit expectation}
\end{equation}
are the bit-width expectations for weights and activations, respectively.
This relaxation enables learning by gradient descent in the space of continuous parameters $\{\alpha, \beta\}$, which is much less expensive than combinatorial search over the configurations of $\{o_\alpha,o_\beta\}$.

\subsection{Efficient Composite Convolution}
\label{subsec:efficient search}

While efficient, differentiable architecture search is not without limitations. A common difficulty for general NAS is the
linear increase in computation and memory with the search space dimension \cite{liu2018darts,DBLP:conf/icml/BenderKZVL18}.
If there are ten candidate operators for a layer, they all need to be applied to the same input, in parallel.
This makes the search impossible for large networks, e.g. the ResNet-50, Inception-V3, etc., and
is usually addressed by resorting to a proxy task.

Unlike general NAS, where the candidate operators are heterogeneous \cite{DBLP:journals/corr/ZophL16,liu2018darts,DBLP:conf/icml/BenderKZVL18}, e.g. convolution, skip connection, pooling, etc., the candidate operators of MPS are homogeneous, namely replicas of the same filter with different bit-widths. This property can be exploited to avoid the expensive parallel operations.
The weighted sum of parallel convolutions, as shown in Figure \ref{fig:mixprecision}, given the weighted activation sum $\bar{a}(x)=\sum_{j=1}^{n_a}\pi^\beta_j a_j(x)$, can be rewritten as
\begin{align}
\label{equ:efficient conv}\nonumber
y=&\sum_{i=1}^{n_f}\pi^\alpha_i f_i\left(\bar{a}(x)\right)
=\sum_{i=1}^{n_f}\pi^\alpha_i \left(Q_i (\mathbf{W}_i)\ast\bar{a}(x)\right) \\
=&\left(\sum_{i=1}^{n_f}\pi^\alpha_i Q_i (\mathbf{W}_i) \right) \ast \bar{a}(x)
= \bar{f}(\bar{a}(x))
\end{align}
where $\bar{f}$ is the composite filter parameterized by weight tensor
\begin{equation}
\mathbf{\overline{W}} = \sum_{i=1}^{n_f}\pi^\alpha_i Q_i (\mathbf{W}_i).
\label{equ:composite}
\end{equation}
Hence, rather
that $n_f$ convolutions between the different $Q_i(\textbf{W}_i)$ and the common activation $\bar{a}(x)$, the architecture of Figure~\ref{fig:mixprecision} only requires one convolution with the composite filter of~(\ref{equ:composite}).  This enables constant training time, independently of the number of bit-widths considered per filter, and the training of large networks become feasible.

In (\ref{equ:composite}), each candidate operator has its own weight tensor $\textbf{W}_i$.
During training, the gradients arriving at a layer are distributed to the different branches. In result, filters of low probability $\pi^\alpha_i$ receive few gradient updates and could be under-trained. For example, a branch of $\pi^\alpha = 0.1$ will only
receive 10\% of the overall gradients. A more robust solution is to share weights across all filters by making $\textbf{W}_i = \textbf{W}$ and redefining the composite filter of (\ref{equ:composite}) as
\begin{equation}
\mathbf{\overline{W}} = \sum_{i=1}^{n_f}\pi^\alpha_i Q_i (\mathbf{W}).
\label{equ:composite_shared}
\end{equation}
In this case, while the gradients are still distributed to each branch, they are all accumulated to update the universal weight tensor $\textbf{W}$. This eliminates the potential for underfitting, and the size of search model can also remain constant independent of the search space. Note that this weight sharing requires a universal $\textbf{W}$ representative enough for multiple quantizers of different bit-widths. While this has not been shown in the low-precision network  literature \cite{DBLP:journals/corr/ZhouNZWWZ16,DBLP:conf/cvpr/CaiHSV17,DBLP:conf/cvpr/ZhuangSTL018,DBLP:conf/eccv/ZhangYYH18}, it is less of a problem for
MPS since 1) the parallel branches learn similar and potentially redundant representations at different bit-widths; 2) what matters for MPS is to optimize the bit allocation, not the weight tensor.

\subsection{Search Space}

The design of the search space is critical for NAS. Unlike NAS, which has an open-set search space, the search space of MPS is well-defined and limited to a relatively small number of possibilities per filter and activation, e.g. bit-widths in $\{1, \ldots, 32\}$. Previous works \cite{wu2018mixed,guo2019single} have coupled weight and activation bit-widths, e.g. defining pairs $(1,4)$, $(2,4)$, etc. This results in $|\cal{B}^\alpha|\times|\cal{B}^\beta|$ parallel branches per filter. To reduce complexity, \cite{wu2018mixed,guo2019single} manually pruned $|\cal{B}^\alpha|\times|\cal{B}^\beta|$ to a small subset, e.g. six pairs, which is suboptimal. Instead, we decouple weight and activation bit-widths, using $|\cal{B}^\alpha|$ and $|\cal{B}^\beta|$ parallel branches for weight and activations, respectively, as shown in Figure \ref{fig:mixprecision}. This decoupling maintains the search space at full size $|\cal{B}^\alpha|\times|\cal{B}^\beta|$, but significantly reduces computation and memory. Since 1) many works \cite{DBLP:conf/eccv/ZhangYYH18,choi2018pact}
have shown that a bit-width of 4 suffices for very good performance; 2) very low-precision networks, e.g. using 2 bits, are the most challenging to develop; and 3) 1-bit activations are usually insufficient for good performance  \cite{DBLP:conf/nips/HubaraCSEB16,DBLP:conf/eccv/RastegariORF16,DBLP:journals/corr/ZhouNZWWZ16,DBLP:conf/cvpr/CaiHSV17}, we use a search space with
\begin{equation}
\label{equ:search space}
\mathcal{B}^\alpha = \{1, 2, 3, 4\}, \quad
\mathcal{B}^\beta = \{2, 3, 4\}.
\end{equation}

\subsection{Learning}
\label{subsec:optimization}

This leads to the learning procedure for MPS. EdMIPS optimizes the Lagrangian of (\ref{equ:lagrangian}) with respect to the architecture parameters $\{\alpha, \beta\}$ of~(\ref{equ:pi}) and the weight tensors $\mathbf{W}$ of ~(\ref{equ:composite_shared}) over the search space of (\ref{equ:search space}).
Both $\textbf{W}$ and
 $\{\alpha, \beta\}$
are learned by gradient descent. This is modeled as a bi-level optimization problem \cite{DBLP:journals/anor/ColsonMS07} in \cite{liu2018darts}. However, this is too complex, even impractical, for searching large models (e.g. ResNet and Inception-V3) on
large datasets (e.g. ImageNet) without a proxy task.
To avoid this, we consider two more efficient optimization approaches. The first, which is used by default, is vanilla end-to-end
backpropagation. It treats the  architecture
$\{\alpha, \beta\}$ and filter  $\textbf{W}$ parameters equally, updating both in a single forward-backward pass. The
second is an alternating optimization of
two steps: 1) fix $\{\alpha, \beta\}$ and update
$\textbf{W}$; 2) fix
$\textbf{W}$ and update $\{\alpha, \beta\}$. It has twice the complexity of vanilla backpropagation.
Our experiments show that these
strategies are efficient and effective for MPS.

\subsection{Architecture Discretization}
\label{subsec:discretization}

Given the optimal architecture parameters $\{\alpha^*,\beta^*\}$, the mixed-precision network must be derived by discretizing the soft selector variables $\pi$ of (\ref{equ:pi}) into the binary selectors $o$ required by (\ref{equ:hard mix}). Two strategies are explored.

The first is a ``winner-take-all'' approach, used  by  default, in which only the branch with the highest $\pi$ is selected and the others removed, by defining
\begin{equation}
\label{equ:argmax}
o^*_i =\left\{
  \begin{array}{cl}
    1, &\quad\textrm{if $i=\arg\max_j \pi_j$,}\\
    0, &\quad\textrm{otherwise.} \end{array}\right.
\end{equation}
This results in a deterministic architecture, which is not affected by the details of the distribution of $\pi$, only the relative ranking of its values.

\begin{figure*}[!t]
\begin{minipage}[b]{.23\linewidth}
\centering
\centerline{\epsfig{figure=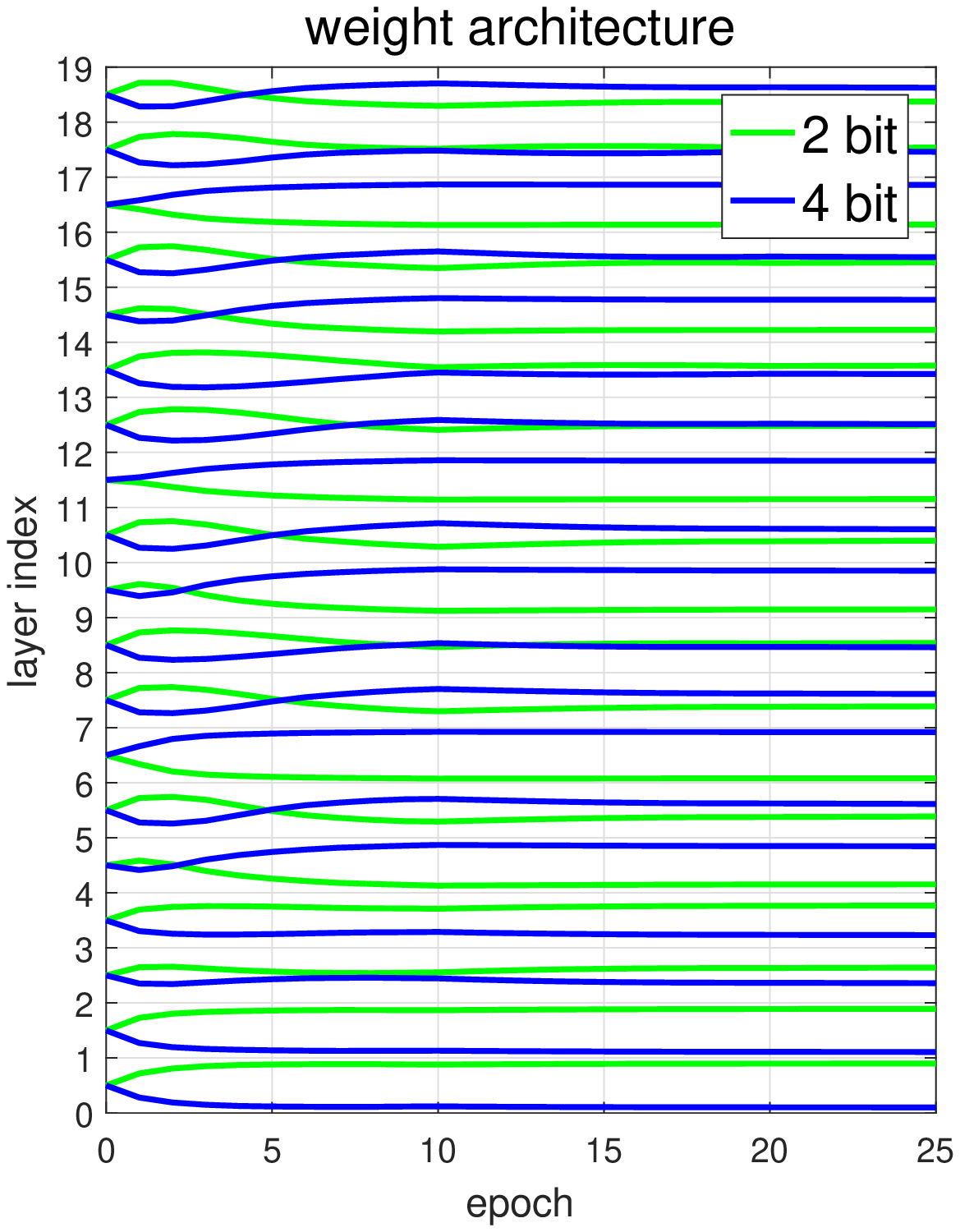,width=4.5cm,height=5.4cm}}
\end{minipage}
\hfill
\begin{minipage}[b]{.23\linewidth}
\centering
\centerline{\epsfig{figure=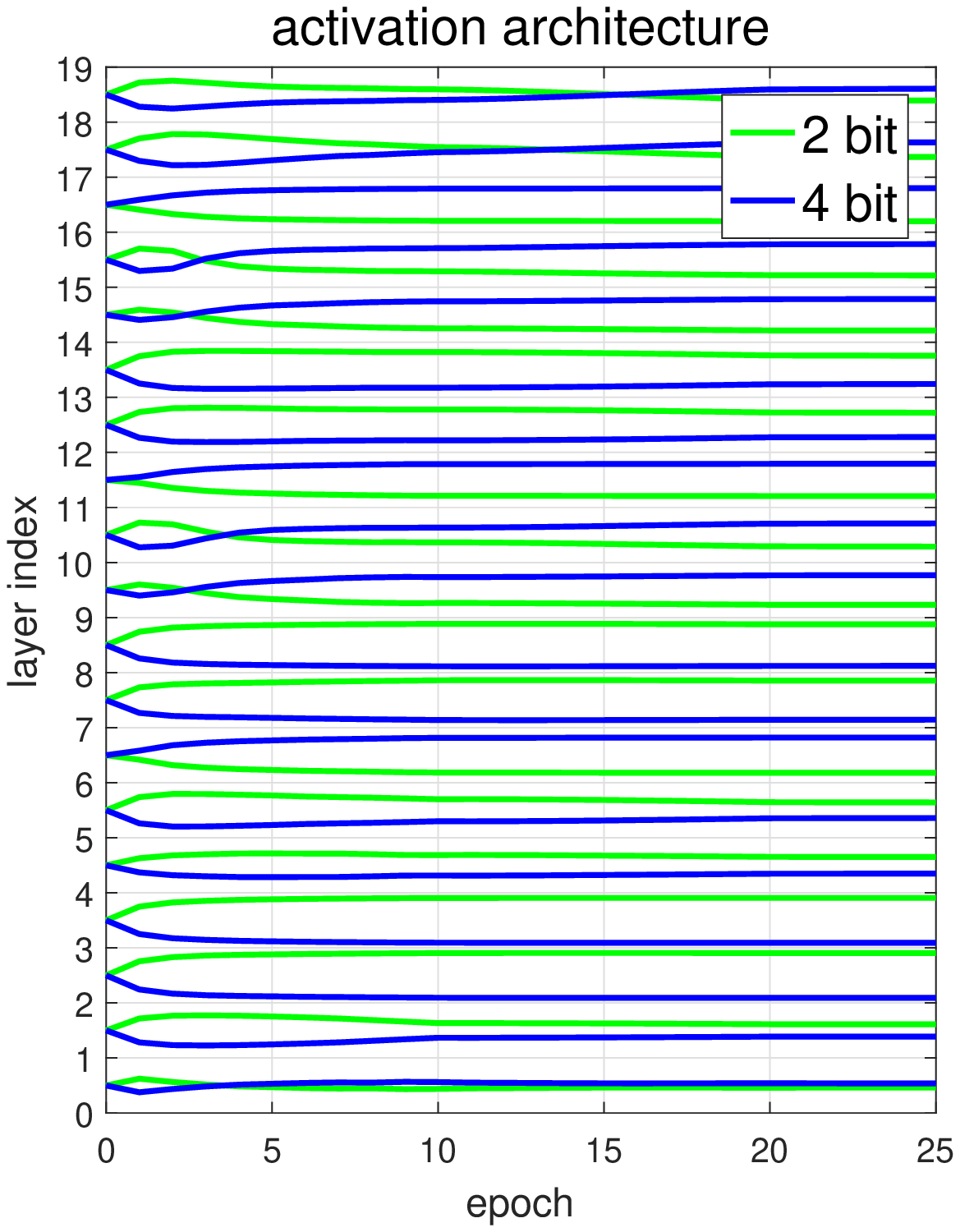,width=4.5cm,height=5.4cm}}
\end{minipage}
\hfill\vline\hfill
\begin{minipage}[b]{.23\linewidth}
\centering
\centerline{\epsfig{figure=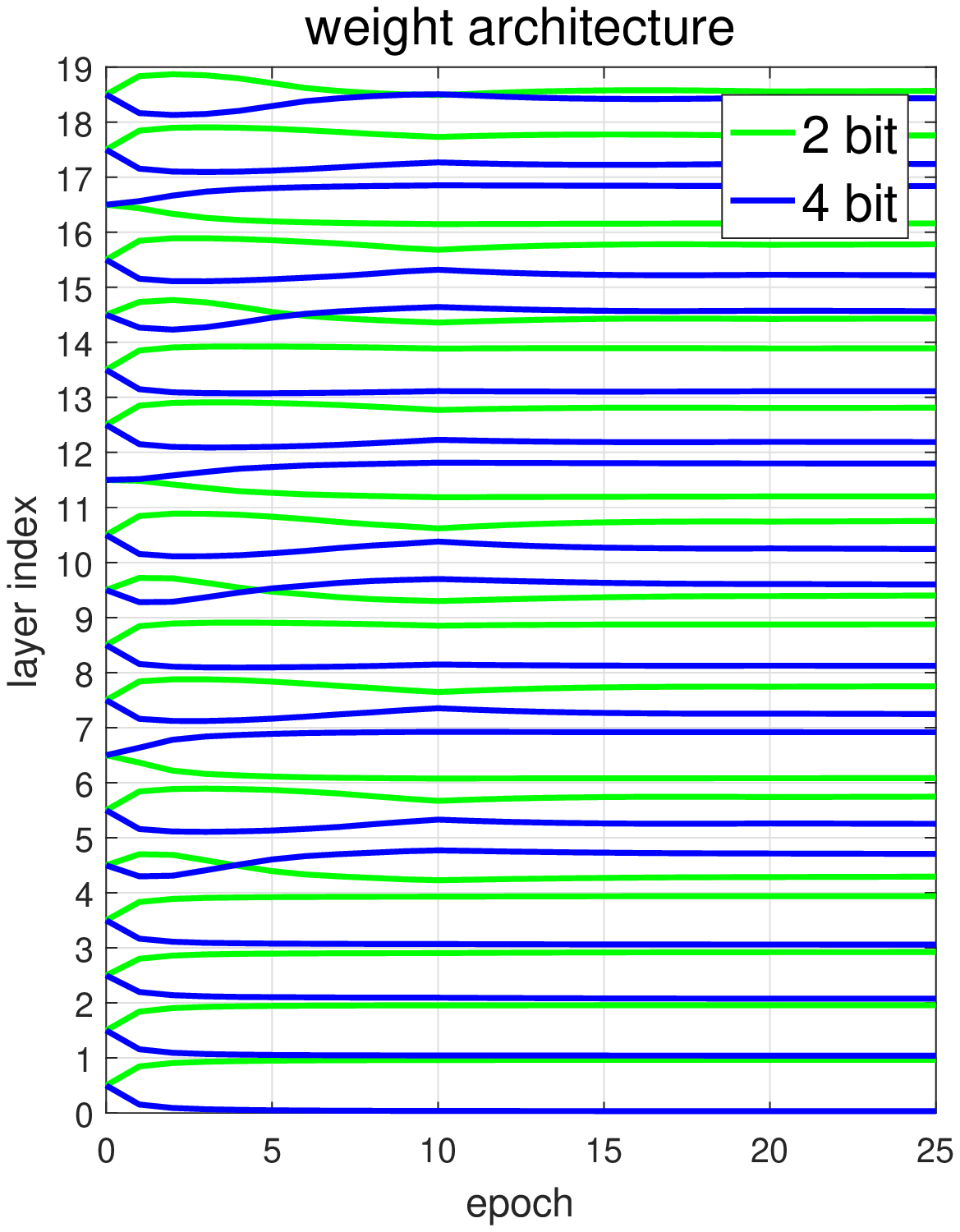,width=4.5cm,height=5.4cm}}
\end{minipage}
\hfill
\begin{minipage}[b]{.23\linewidth}
\centering
\centerline{\epsfig{figure=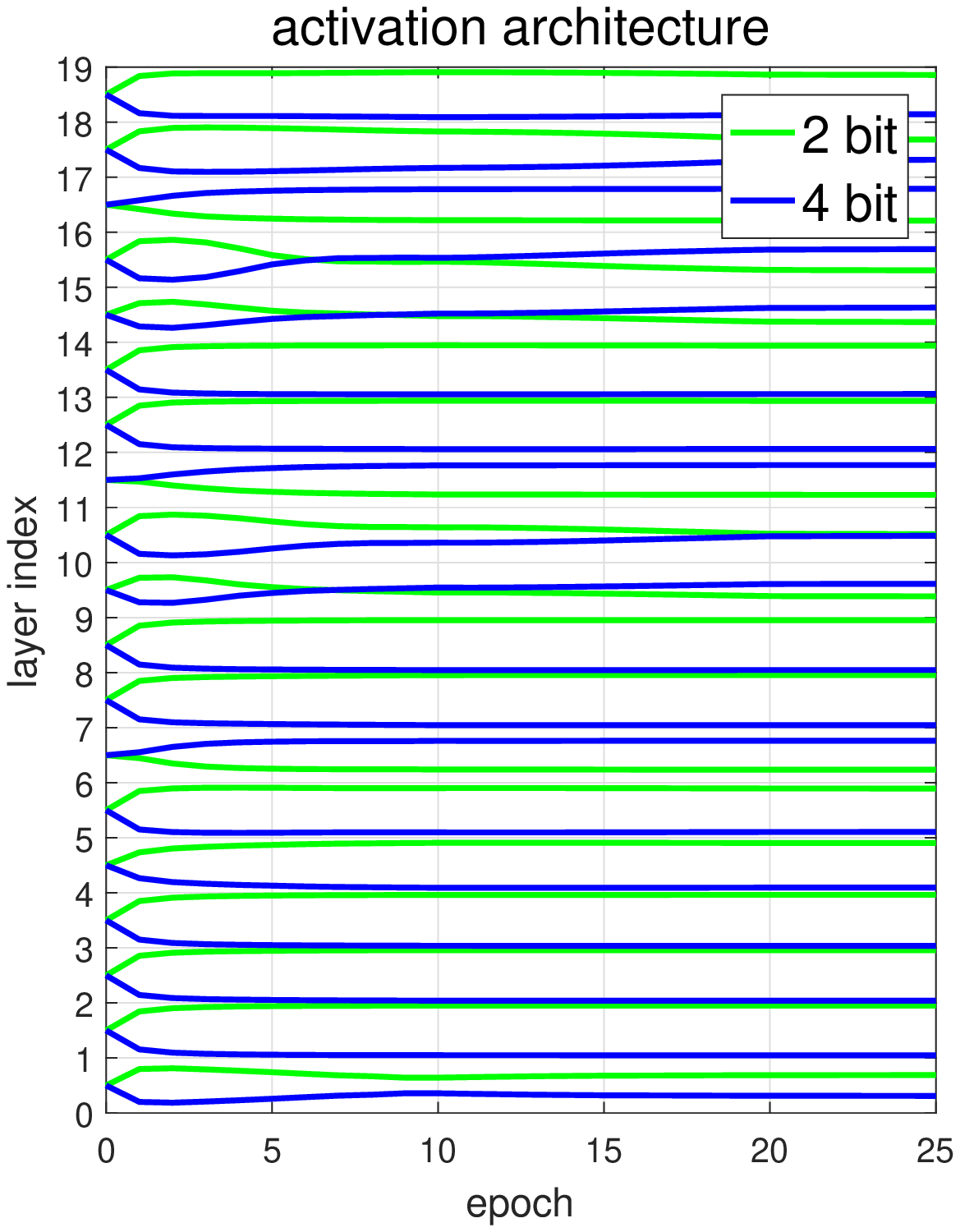,width=4.5cm,height=5.4cm}}
\end{minipage}
\caption{Architecture evolution of ResNet-18 during search. The curves shown per layer index represent the probabilities of the two bit-widths. Layers indexes are as in PyTorch, where 7/12/17 are residual connections. Left: $\eta=0.001$ in (\ref{equ:lagrangian}). Right: $\eta=0.002$.}
\label{fig:evolution}
\end{figure*}

The second is a sampling strategy. Since each $\pi$ is a categorical distribution, hard selectors $o_i$ can be  sampled from a multinomial distribution
\begin{equation}
\label{equ:multinomial}
O\sim\mathbf{Multinomial}(n,\pi).
\end{equation}
with one trial $n=1$.
This defines a random architecture with expected BitOps as defined in (\ref{equ:bit expectation}). The user can then sample multiple architectures and choose the one with the best trade-off between accuracy and complexity, as in \cite{wu2018mixed}. However, in our experience, the variance of this random architecture is very large, making it very inefficient to find the best network configuration. Alternatively, it is possible to sample from (\ref{equ:multinomial}) with multiple trials, e.g. $n=50$, and choose the architecture with the highest counts. This produces architectures closer to that  discovered by ``winner-take-all'', converging to the latter when $n\rightarrow\infty$.

\section{Experiments}

EdMIPS was evaluated with ImageNet \cite{DBLP:journals/ijcv/RussakovskyDSKS15},
top-1 and top-5 classification accuracy. Previous MPS works
\cite{wu2018mixed,wang2019haq} used ResNet-18 or MobileNet
\cite{DBLP:journals/corr/HowardZCKWWAA17}. However, these are relatively simple networks, that stack replicas of a single module block, and for which heuristics can be used to hand-craft nearly optimal bit allocation. For example, the depth-wise layers of MobileNet should receive more bits than the point-wise ones. On the other end of the spectrum, GoogLeNet \cite{DBLP:conf/cvpr/SzegedyLJSRAEVR15} and Inception-V3 \cite{DBLP:conf/cvpr/SzegedyVISW16} are complex combinations of different modules and too complicated to optimize by hand. Since this is the case where MPS is more useful, beyond the small and simple AlexNet and ResNet-18, we have also tested EdMIPS on larger and more complicated ResNet-50, GoogLeNet and Inception-V3. These experiments
establish broad baselines for future studies in MPS.

\subsection{Implementation Details}
\label{subsec:details}

All experiments, followed standard ImageNet training on PyTorch\footnote{https://github.com/pytorch/pytorch}, with the following exceptions. For simplicity, all auxiliary losses were removed from the GoogLeNet and Inception-V3.
EdMIPS used a learning rate of 0.1 for network parameters $\textbf{W}$, and 0.01 for
architecture parameters $\{\alpha,\beta\}$. All network parameters were initialized as usual,
and all architecture parameters were set to 0.01, treating all candidates equally. The search
model was trained for 25 epochs, with learning rates decayed by 10 time at every 10 epoches.
After training the search model, the classification model was derived by the ``winner-take-all''
strategy of Section \ref{subsec:discretization}. Next, the classification models were trained
for 50 (95) epochs, with learning rate decayed by 10 times at every 15 (30) epochs, to allow
exploration (the final comparison with the state-of-the-art).
All models were trained from scratch. For network quantization, we followed \cite{DBLP:conf/cvpr/CaiHSV17} but enabled scaling in all BatchNorm layers.

\begin{figure*}[!t]
\begin{minipage}[b]{.19\linewidth}
\centering
\centerline{\epsfig{figure=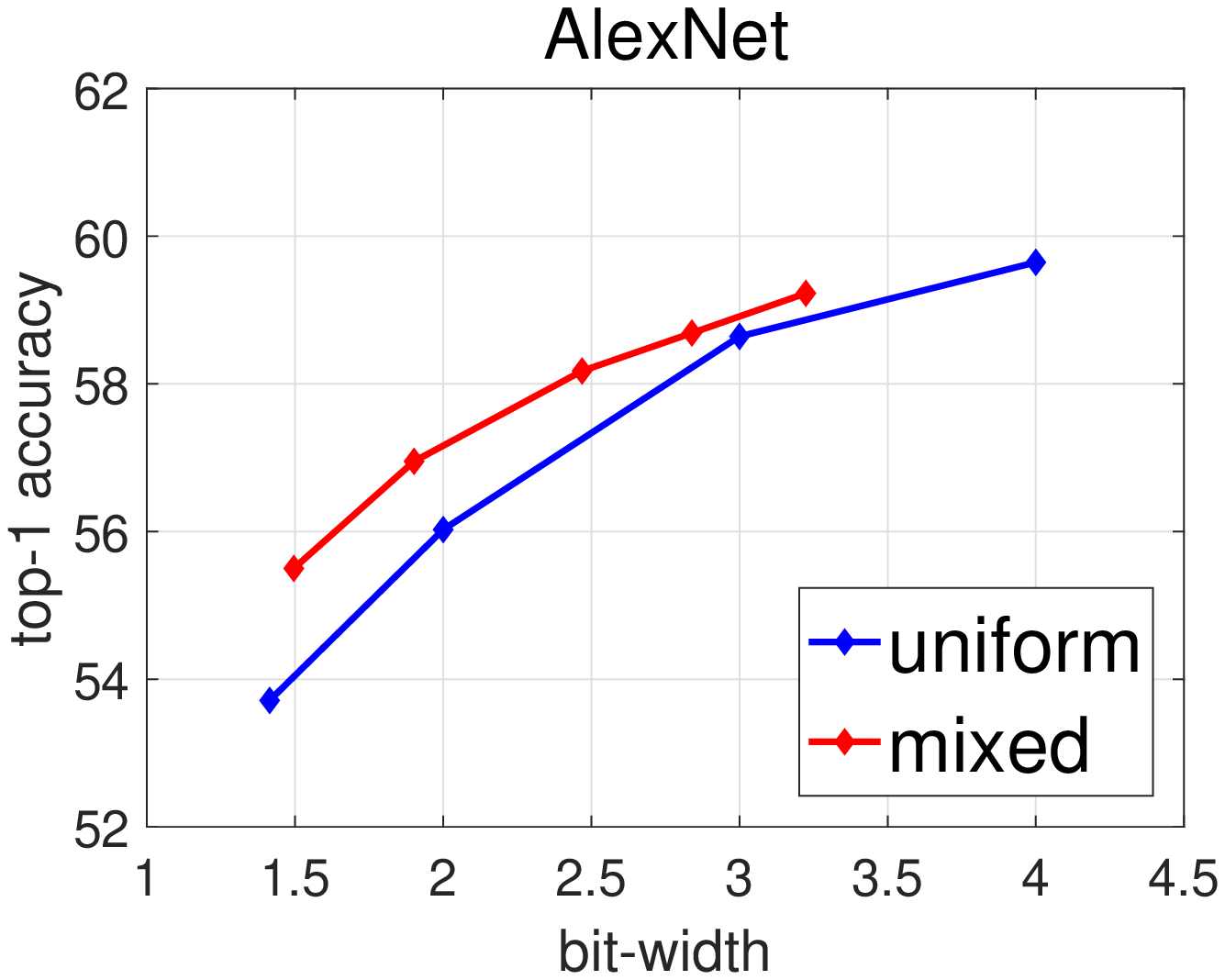,width=3.6cm,height=2.7cm}}
\end{minipage}
\hfill
\begin{minipage}[b]{.19\linewidth}
\centering
\centerline{\epsfig{figure=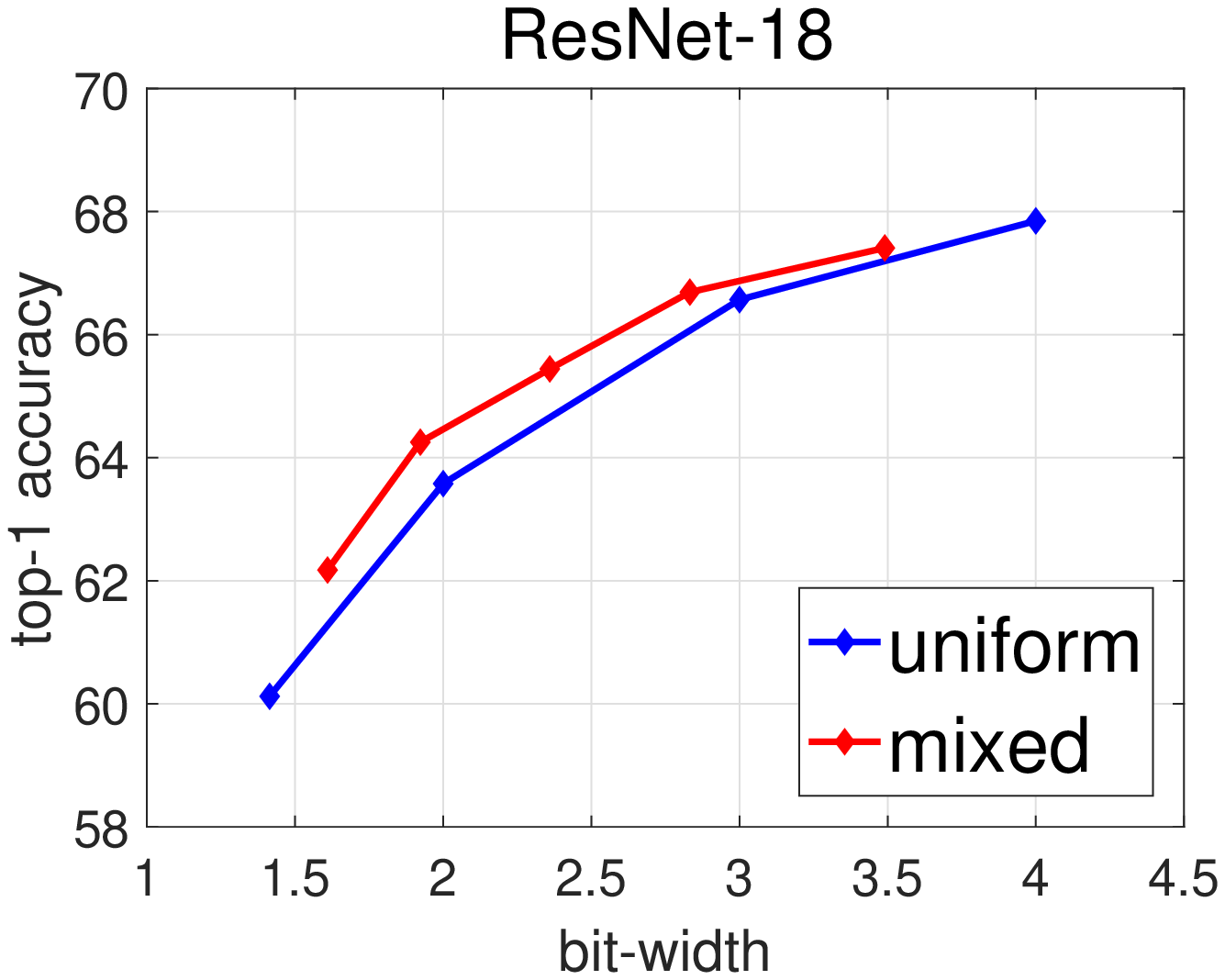,width=3.6cm,height=2.7cm}}
\end{minipage}
\hfill
\begin{minipage}[b]{.19\linewidth}
\centering
\centerline{\epsfig{figure=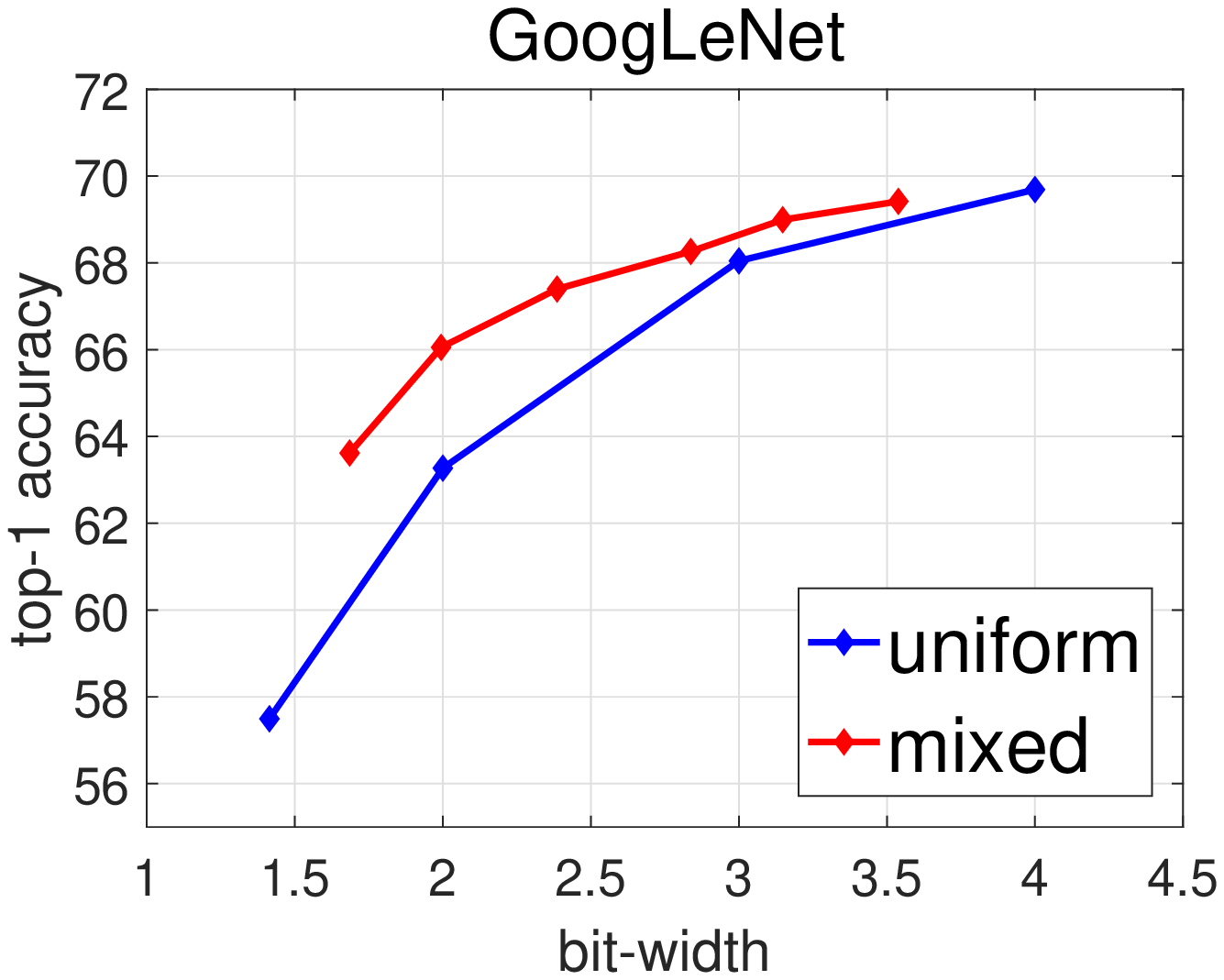,width=3.6cm,height=2.7cm}}
\end{minipage}
\hfill
\begin{minipage}[b]{.19\linewidth}
\centering
\centerline{\epsfig{figure=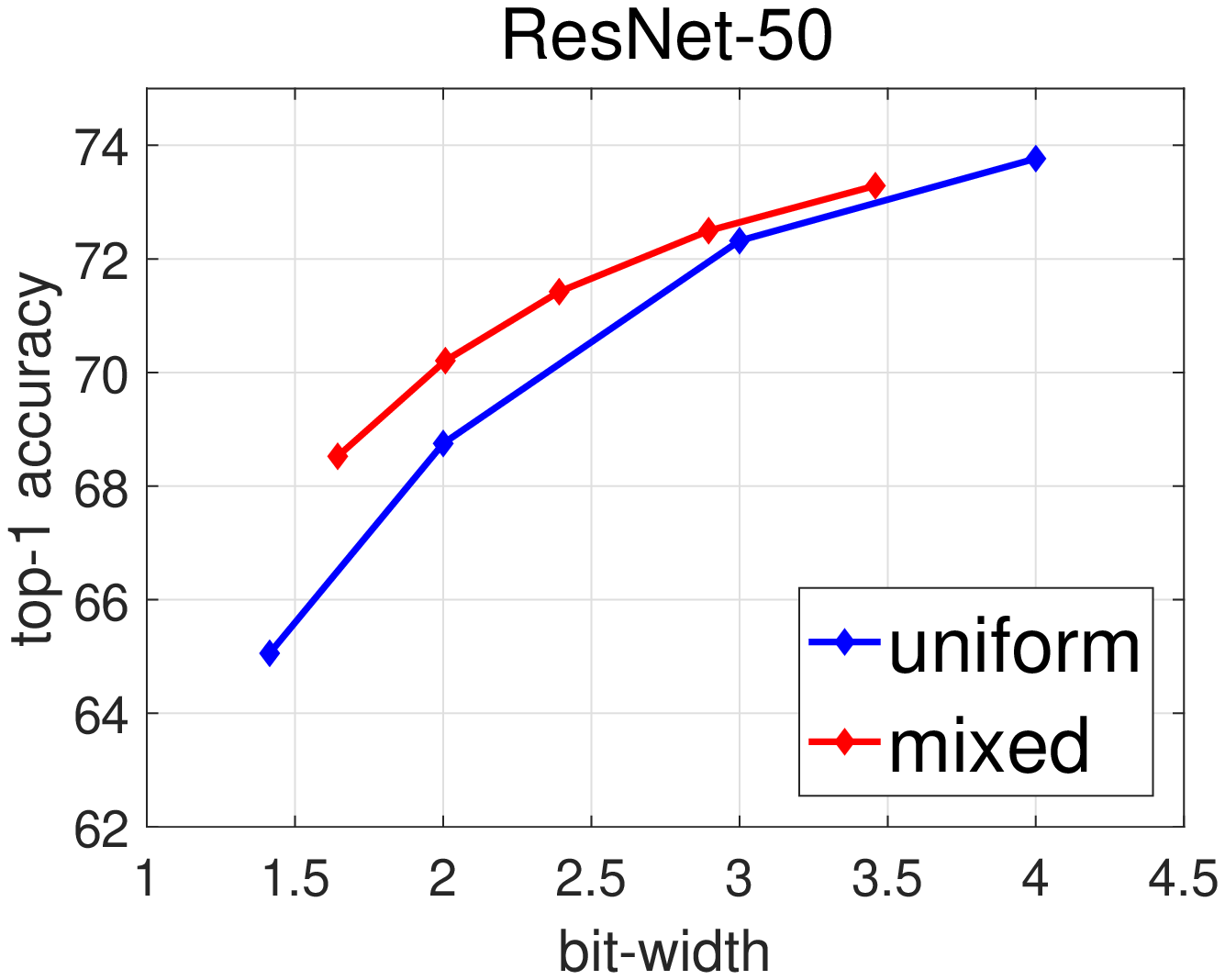,width=3.6cm,height=2.7cm}}
\end{minipage}
\hfill
\begin{minipage}[b]{.19\linewidth}
\centering
\centerline{\epsfig{figure=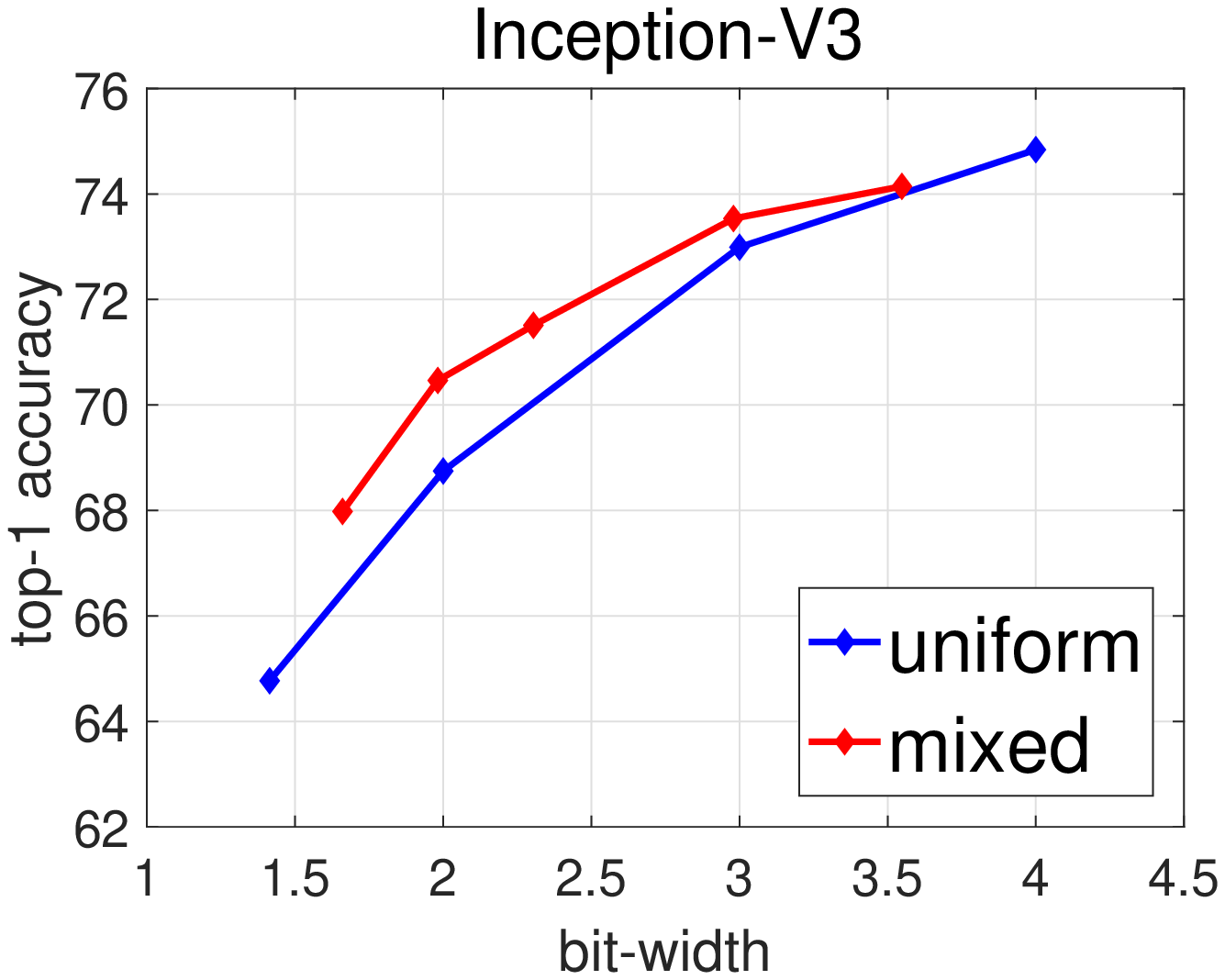,width=3.6cm,height=2.7cm}}
\end{minipage}
\caption{Comparison of the uniform HWGQ-Net and the EdMIPS network. The x-axis, indicating BitOps, is normalized to the scale of bit-width, which is actually in log-scale.}
\label{fig:accuracy}
\end{figure*}

\begin{figure*}[!t]
\begin{minipage}[b]{0.104\linewidth}
\centering
\centerline{\epsfig{figure=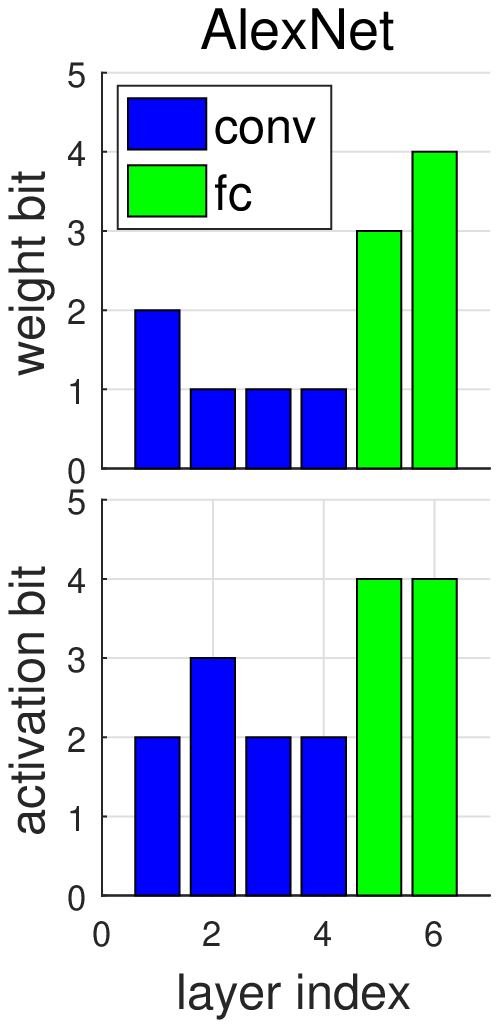,width=1.96cm,height=3.38cm}}
\end{minipage}
\hfill
\begin{minipage}[b]{0.23\linewidth}
\centering
\centerline{\epsfig{figure=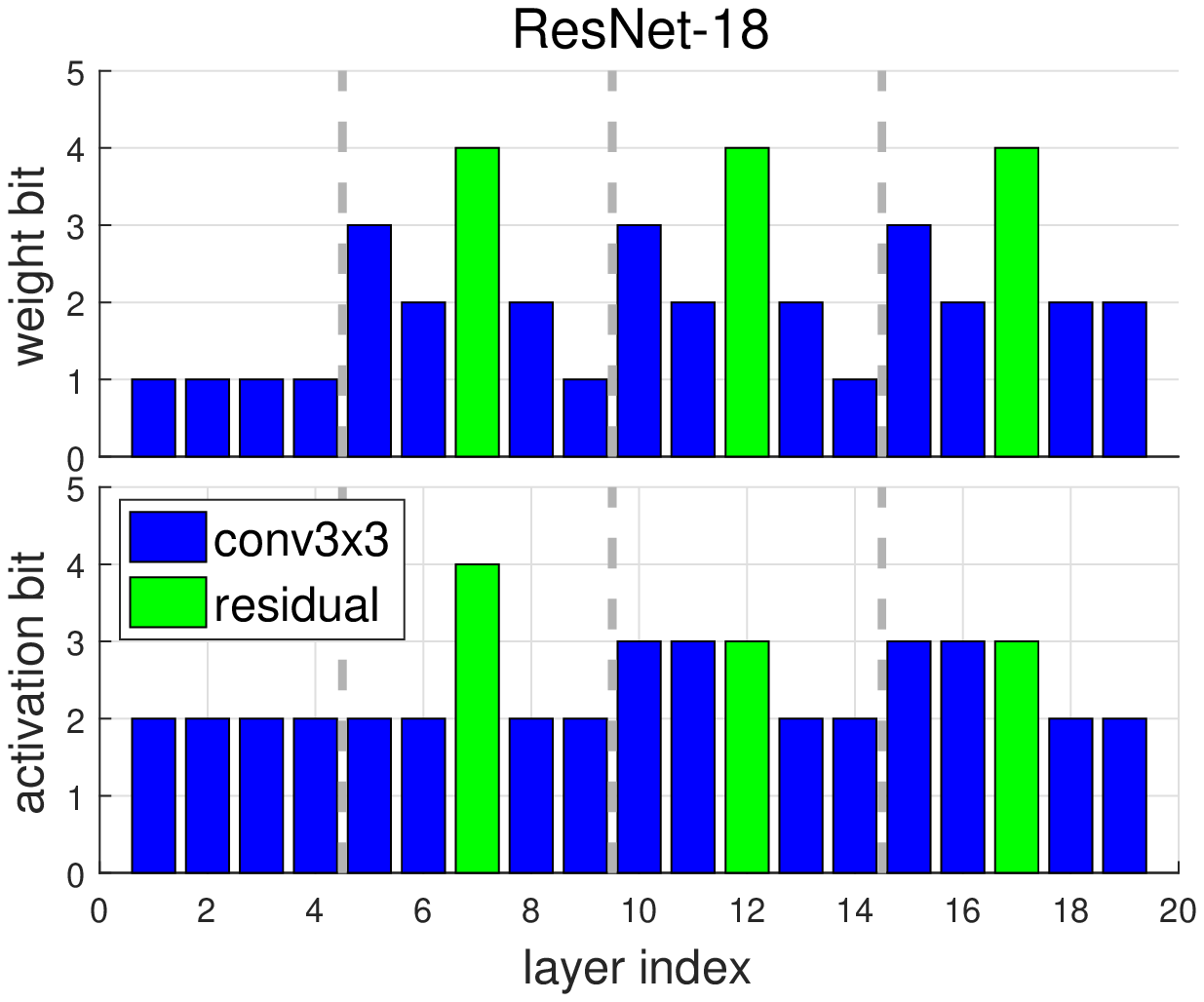,width=4.19cm,height=3.38cm}}
\end{minipage}
\begin{minipage}[b]{0.65\linewidth}
\centering
\centerline{\epsfig{figure=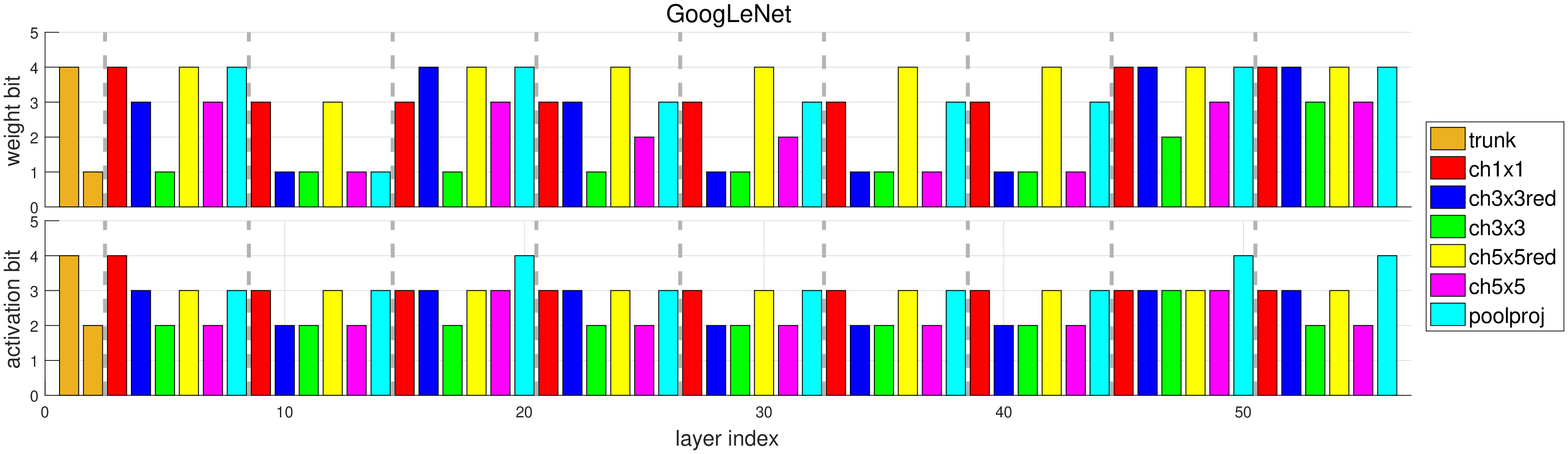,width=11.8cm,height=3.38cm}}
\end{minipage}\\
\begin{minipage}[h]{0.99\linewidth}
\centering
\centerline{\epsfig{figure=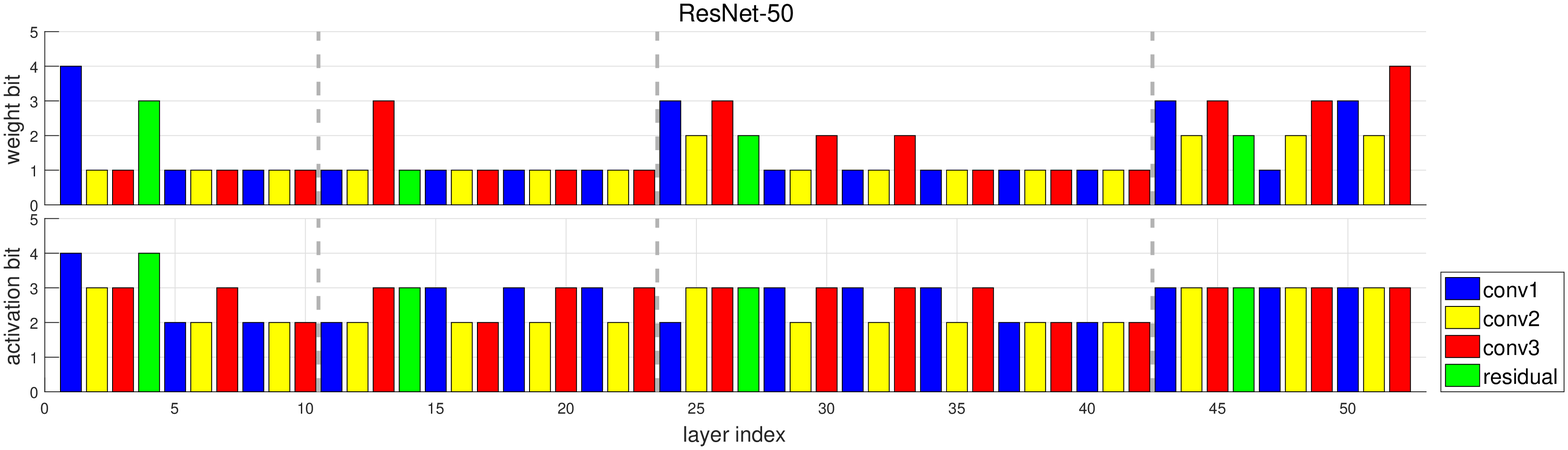,width=9.98cm,height=3.38cm}}
\end{minipage}\\
\hfill
\begin{minipage}[h]{0.99\linewidth}
\centering
\centerline{\epsfig{figure=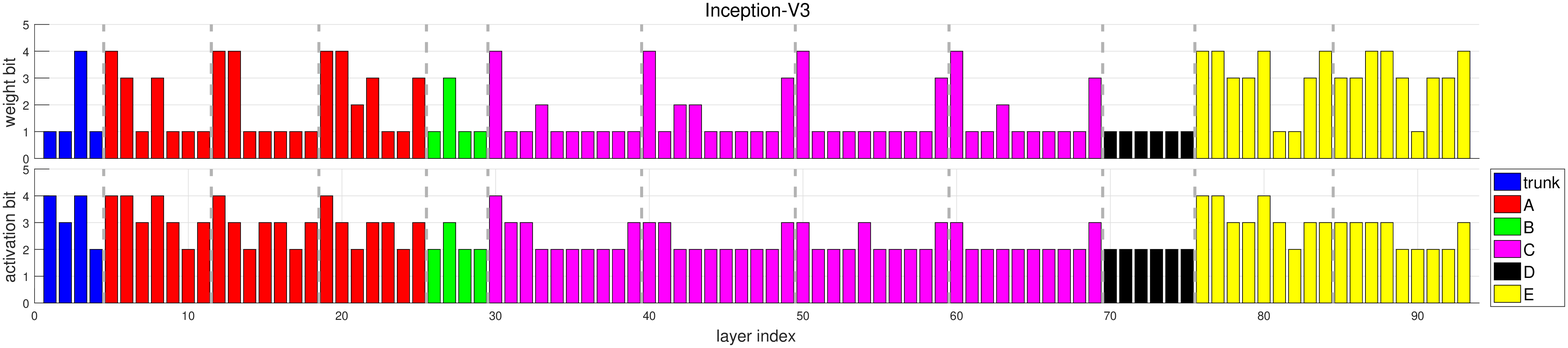,width=15.2cm,height=3.38cm}}
\end{minipage}
\caption{EdMIPS bit allocation for AlexNet, ResNet-18, GoogLeNet, ResNet-50 and Inception-V3.}
\label{fig:allocation}
\end{figure*}

\subsection{Architecture Evolution}

Sometimes, it is mysterious why and how an architecture is found by NAS. Due to the
differentiability of EdMIPS, the soft architecture evolves gradually during search, as shown in Figure \ref{fig:evolution}. For clarity, we only show the evolution of a ResNet-18 with search space of $\{2, 4\}$ bit. Some interesting observations follow from the left part of Figure \ref{fig:evolution}. First, the use of (\ref{equ:lagrangian}) with complexity penalty $\eta=0.001$ is shown to avoid the trivial solution (highest bit-width always selected). Candidates of lower bit-width are selected for many layers.
Second, nearly all layers choose 2 bits in the early epochs, where the network parameters are not yet trained and the complexity penalty dominates. As network parameters become stronger, the error penalty is more likely to overcome the complexity penalty, and 4-bit candidates start to be selected.
Third, weights and activations can have different optimal bit allocations. For example, 4-bits are
selected for the weights of layers 5/6/8, whose activations are assigned 2-bits. Fourth, since the complexity penalty of residual connections (layers 7/12/17) is much smaller than 3$\times$3 convolutions, they are allocated 4-bits early on. Finally, both candidates are equally strong for some layers, e.g. 9/13 for weights and 1 for activations. In these cases, the model is not confident on which
candidate to choose.

\subsection{Effect of Complexity Constraint}

When a stronger complexity constraint is enforced, lower bit-widths are more likely to be chosen. This can be observed by comparing the left ($\eta=0.001$) and right ($\eta=0.002$) of Figure \ref{fig:evolution}. For layers preferring lower bit-width, this preference is reinforced by stronger $\eta$. For example, in layers 1-4 for weights and 2-6 for activations, EdMIPS converges to lower bit-widths much faster. For layers preferring lower bit-width at the start but switching to more bits at the end, e.g. 5/10/15 for weights and 10/15/16 for activations, more epochs are required to reach the crossing point. The stronger complexity constraint can also change the final decisions. For example, for weights at layers 6/8/11 and activations at layers 18-19, the decisions switch from 4- to 2-bit. These results show that the optimal architecture depends on the complexity constraint.

\subsection{Comparison to Uniform Bit Allocation}
\label{subsec:comprison with uniform}

Figure \ref{fig:accuracy} shows that EdMIPS networks substantially outperform the uniform HWGQ-Net versions of AlexNet, ResNet-18/50, GoogLeNet, and Inception-V3. Note that the HWGQ-Net has fairly high baselines. Since the learned model is bounded by the weakest (W1A2) and strongest (W4A4) models in the search space, the improvements are small on both ends, but substantial in between. For example, the EdMIPS models  improve the uniform 2-bit HWGQ-Net by about 0.9 point for AlexNet, 0.8 point for ResNet-18, 2.8 points for GoogLeNet, 1.5 points for ResNet-50 and 1.7 points for Inception-V3. This is strong evidence for the effectiveness of EdMIPS.

\begin{figure*}[!t]
\begin{minipage}[b]{.33\linewidth}
\centering
\centerline{\epsfig{figure=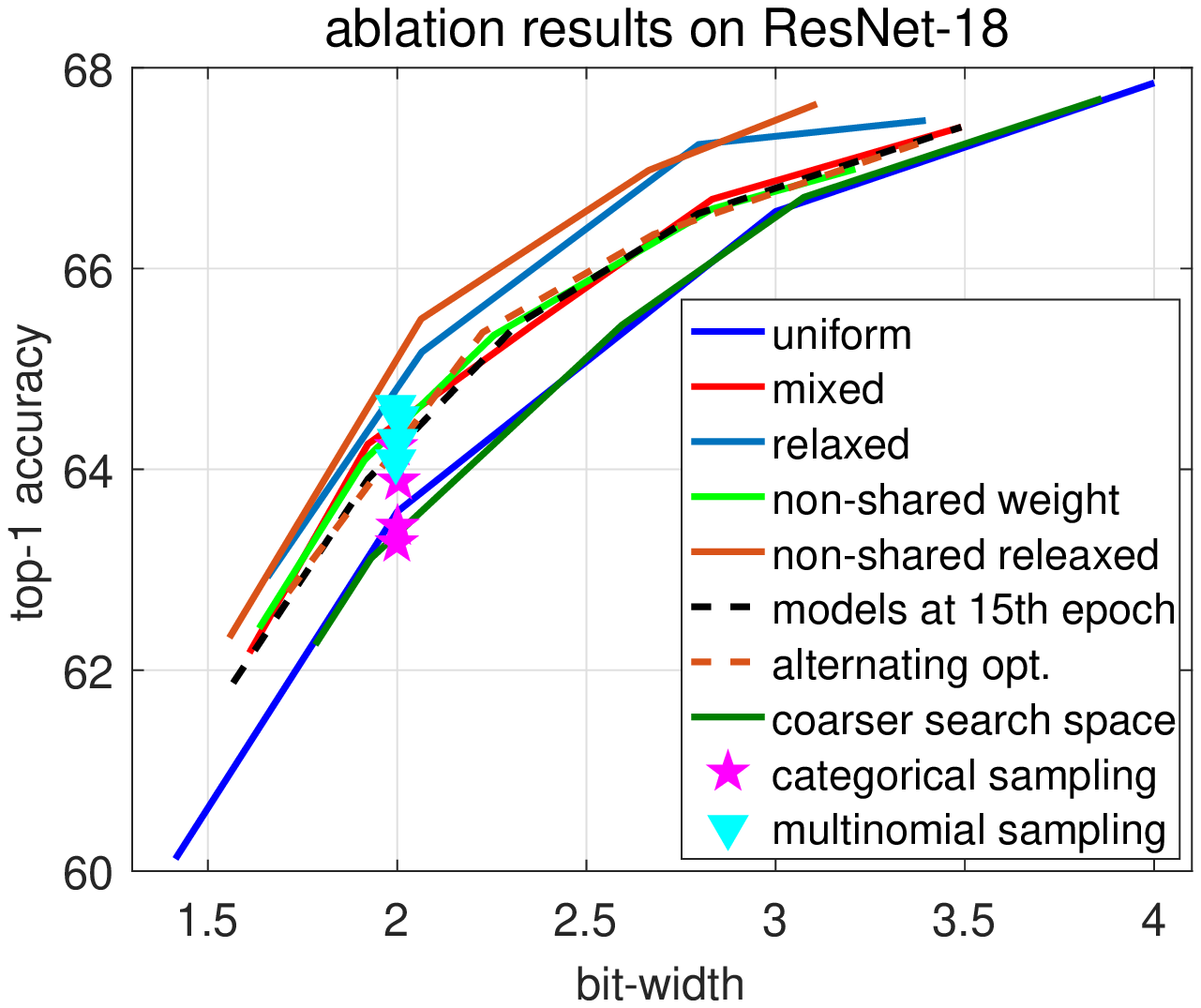,width=6cm,height=4.5cm}}{(a)}
\end{minipage}
\hfill
\begin{minipage}[b]{.33\linewidth}
\centering
\centerline{\epsfig{figure=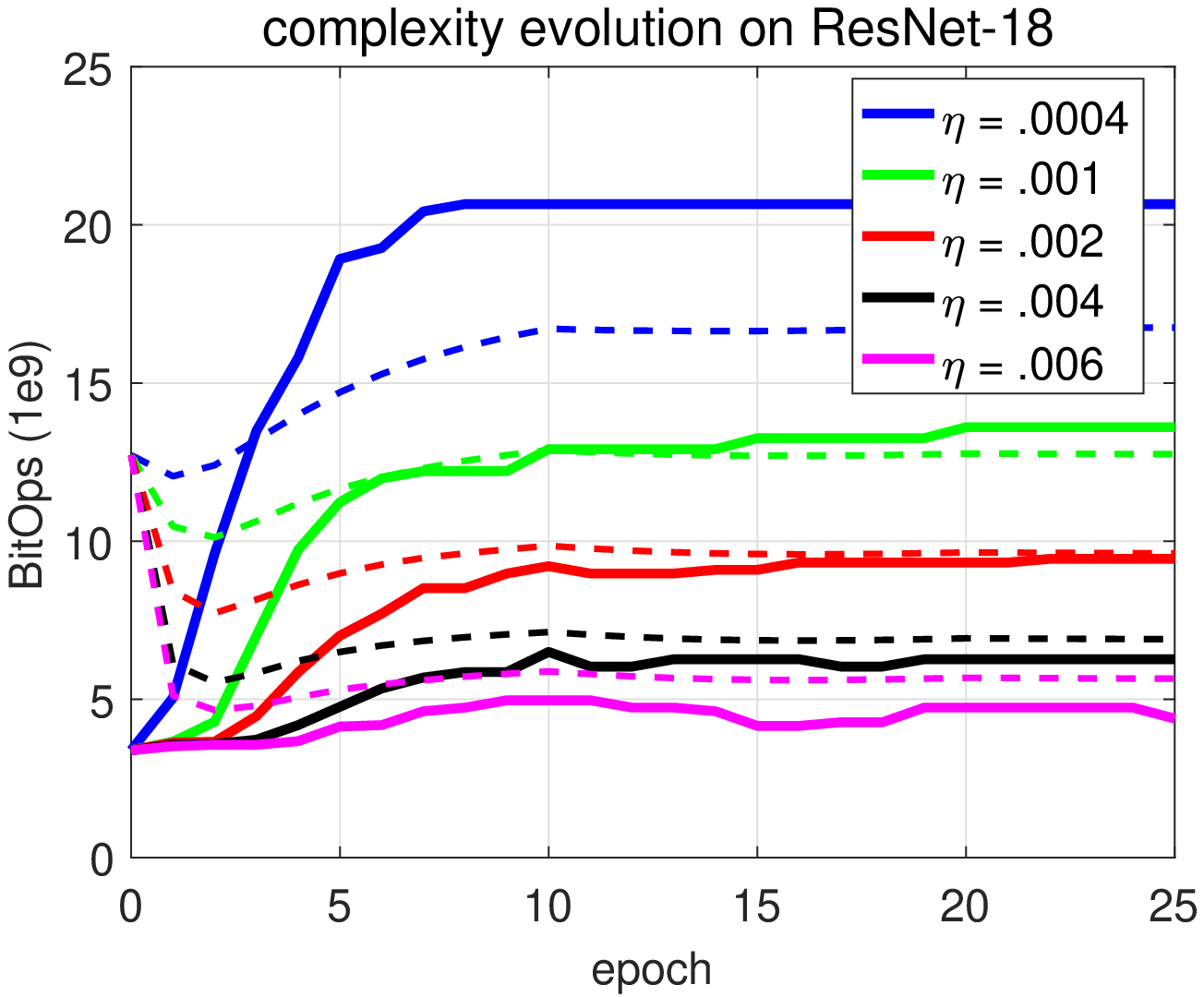,width=6cm,height=4.5cm}}{(b)}
\end{minipage}
\hfill
\begin{minipage}[b]{.33\linewidth}
\centering
\centerline{\epsfig{figure=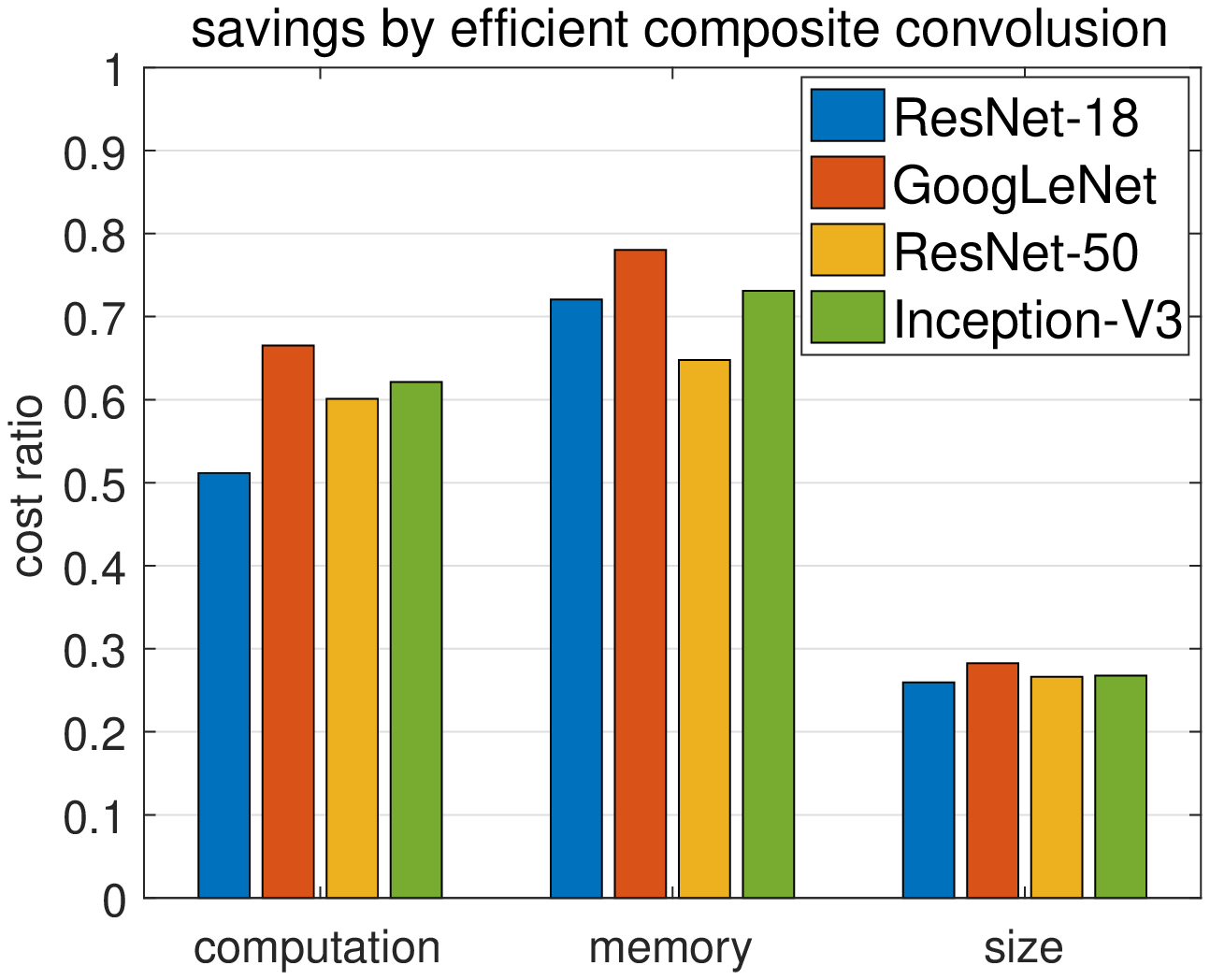,width=6cm,height=4.5cm}}{(c)}
\end{minipage}
\caption{(a) ablation experiments on ResNet-18; (b) complexity evolution on ResNet-18 (dashed lines are expected complexities of (\ref{equ:bit expectation})); (c) savings by efficient composite convolution of Section \ref{subsec:efficient search}.}
\label{fig:ablation}
\end{figure*}

\subsection{Learned Optimal Bit Allocation}

To understand what EdMIPS learns, we visualize the optimal bit allocation of each layer in
Figure \ref{fig:allocation}, for AlexNet, ResNet-18, GoogLeNet, ResNet-50 and Inception-V3
with roughly the same BitOps as the uniform 2-bit models. On AlexNet, whose FC layers are lighter and closer to the output, they receive higher bit-width. The layers closer to the input, e.g. 1st for weights and
2nd for activations, also receive relatively higher bit-width. On ResNet-18, the cheap residual
connections usually receive the largest bit-width. The first block of layers in each stage,
e.g. 5-6th, 10-11th, and 15-16th, also has relatively higher bit-width.
GoogLeNet is a more complicated network, and so is its bit allocation. The most expensive
``ch3x3'' layers frequently receive the lowest bit-width, but not always. Higher bit-widths
are allocated to ``ch3x3'' layers closer to the output. On the other hand, ``ch5x5red'' is
very cheap, and usually receives the highest bit-width. Although ``ch1x1'' is the second most
expensive filter, it is frequently allocated higher bit-width. This reflects its highest
sensitivity, shown in Figure \ref{fig:sensitivity}. Finally, ``ch3x3red'' and ``ch5x5'' have
similar computation, but the former is much more sensitive. Hence, it receives higher
bit-width allocations.

Since the filters in ResNet-50 have relatively close computations due to
its bottleneck design, its bit allocation is closer to uniform than the other networks. However, the layers closer to the inputs and
outputs prefer higher bit-widths, and ``conv1'' and ``conv3'' receive higher bit-widths than ``conv2'' in the bottleneck module. Inception-V3 is a very complicated architecture, with 93 filters in total to search. Module-D chooses the lowest bit-width, although it does not have very high computations. In
module-A and module-C, higher bit-widths are allocated to the cheap ``1x1'' filters more often than the other ``3x3'', ``5x5'', ``1x7'' and ``7x1'' filters. Since module-E is
closer to the outputs, many of its filters receive higher bit-width. Note the difficulty of hand-crafting the optimal bit allocations for these complex networks.

\subsection{Ablation Studies}

ResNet-18 is used for ablation experiments. The default EdMIPS model is denoted as ``mixed'' in Figure \ref{fig:ablation} (a).

\noindent{\bf Relaxation:}
Figure \ref{fig:ablation} (a) shows that the relaxed models of (\ref{equ:soft mix}) (trained for 50 epochs), only have slightly higher accuracy than their discretized counterparts, i.e. architecture discretization has little performance cost.

\vspace{0.1cm}
\noindent{\bf Filter Sharing vs. Non-Sharing:}
In the default model, all parallel candidates share the weight tensor of (\ref{equ:composite_shared}). Figure \ref{fig:ablation} (a) shows that this is as effective as learning the individual weight tensors of (\ref{equ:composite}). This is also confirmed by the comparison of their relaxed counterparts, indicating that the unshared parallel branches are somewhat redundant.

\vspace{0.1cm}
\noindent{\bf Sampling:}
Four architectures were sampled from the architecture distribution of the model close to 2-bit of Figure \ref{fig:accuracy}, using both the categorical and multinomial sampling discussed in Section \ref{subsec:discretization}, with close BitOps to the uniform 2-bit model. Figure \ref{fig:ablation} (a) shows that categorical sampling architectures have large accuracy variance, sometimes even underperforming the uniform baseline. Better performance and lower variance are obtained with multinomial sampling.

\vspace{0.1cm}
\noindent{\bf Learning Strategy:}
Figure \ref{fig:ablation} (a) shows that there is little difference in accuracy between the default model, which uses single pass optimization, and a model trained with the alternating optimization of Section \ref{subsec:optimization}. Since the latter has twice the learning complexity, single pass optimization is used by default on EdMIPS.

\vspace{0.1cm}
\noindent{\bf Convergence Speed:}
Figure \ref{fig:ablation} (b)
summarizes  the evolution of network complexity as a function of $\eta$, for the five ResNet-18 models of Figure \ref{fig:accuracy}. The search usually converges quickly, with minor changes in architecture complexity after the 15th epoch. The accuracy of the corresponding classification models (``models at 15th epoch'') is shown in Figure \ref{fig:ablation} (a). They are comparable to those found after 25 search epochs.

\vspace{0.1cm}
\noindent{\bf Search Space:}
Figure \ref{fig:ablation} (a) shows that no gain occurs over the uniform model when a coarser search space ($\{1, 4\}$ bit for weight and $\{2, 4\}$ bit for activation) is used.

\vspace{0.1cm}
\noindent{\bf Efficient Composite Convolution:}
Figure \ref{fig:ablation} (c) summarizes the practical savings achieved with the efficient composite convolution of Section \ref{subsec:efficient search}, with respect to the vanilla parallel convolutions, in terms of computation, memory and model size, for the search space of (\ref{equ:search space}). The weight sharing of (\ref{equ:composite_shared}) reduces model size by a factor of almost four. Replacing the parallel convolutions results in 30-50\% savings in computation and 20-40\% in memory. These savings make MPS much more practical. For example, the time needed to search for the ResNet-18 on 2 GPUs decreased from 35 to 18 hours, and the search for the Inception-V3 can be performed with 8 GPUs and 12GB of memory. In result, the complete search (e.g. 25 epochs) only increases by $\sim$45\% the training complexity of a uniform low-precision ResNet-18 network. It should be mentioned that these practical savings are nevertheless smaller than theoretically predicted by (\ref{equ:efficient conv}), due to other practical cost bottlenecks, e.g. weight/activation quantization, parallelization efficiency in GPUs, network architecture, etc. Better implementations of EdMIPS could enable further savings.

\begin{table}[t]
\tablestyle{1.8pt}{1.2}
\vspace{-0.2cm} \caption{Comparison to the state-of-the-art.}
\label{tab:final}
\begin{tabular}
{c|c V{2.5} x{22}|x{22}|c|c|c}\hline
\multicolumn{2}{c V{2.5}}{Model} &Ref &Full &HWGQ &LQ \cite{DBLP:conf/eccv/ZhangYYH18} &EdMIPS\\
\multicolumn{2}{c V{2.5}}{bit-width} &32 &32 &2 &2 &$\sim$2\\[.1em]\shline
\multirow{2}{*}{AlexNet} &Top-1 &57.1 &59.2  &58.6 &57.4 &59.1\\
&Top-5   &80.2 &81.7  &80.9 &80.1 &81.0\\\hline
\multirow{2}{*}{ResNet-18} &Top-1  &69.6 &70.2  &65.1 &64.9 &65.9\\
&Top-5 &89.2 &89.5  &86.2 &85.9 &86.5\\\hline
\multirow{2}{*}{GoogLeNet} &Top-1  &73.3 &72.7  &64.8 &- &67.8\\
&Top-5 &91.3 &91.0  &86.3 &- &88.0\\\hline
\multirow{2}{*}{ResNet-50} &Top-1  &76.0 &76.2  &70.6 &71.5 &72.1\\
&Top-5 &93.0 &93.0  &89.8 &90.3 &90.6\\\hline
\multirow{2}{*}{Inception-V3} &Top-1  &77.5 &77.3  &71.0 &- &72.4\\
&Top-5 &93.6 &93.6  &89.9 &- &90.7\\\hline
\end{tabular}
\end{table}

\subsection{Comparison to state-of-the-art}

Table \ref{tab:final} compares EdMIPS models to state-of-the-art uniform low-precision networks, including HWGQ-Net \cite{DBLP:conf/cvpr/CaiHSV17} and LQ-Net \cite{DBLP:conf/eccv/ZhangYYH18}.
EdMIPS models, with similar BitOps to the 2-bit uniform models, achieve top performance for all base networks. Compared to the uniform HWGQ-Net, the gains are particularly large (3 points) for the complicated GoogLeNet, a compact model very averse to quantization. On the other hand,
for the simplest model (AlexNet), the 2-bit EdMIPS model is already equivalent to the full
precision network. These results show that EdMIPS is an effective MPS solution.
Since 1) \cite{wang2019haq} only experiments on MobileNet and uses latency as complexity indicator; 2) \cite{wu2018mixed} mainly focuses on ResNet-18/34 of higher bit-widths, e.g. 4-bit; 3) neither of these works have released the code needed to reproduce their results, we are unable to compare to them. We believe that the release of EdMIPS code and the solid baselines now established by Table \ref{tab:final} will enable more extensive comparisons to future work in MPS.

\section{Conclusion}

We have proposed EdMIPS, an efficient framework for MPS based on a differentiable architecture. EdMIPS has multiple novel contributions to the MPS problem. It can search large models, e.g. ResNet-50 and Inception-V3, directly on ImageNet, at affordable costs. The learned mixed-precision models of multiple popular networks substantially outperform their uniform low-precision counterparts and
establish a solid set of baselines for future MPS research.

\paragraph{Acknowledgment} This work was partially funded by NSF awards IIS-1637941, IIS-1924937, and NVIDIA GPU donations.

%

{\small
\bibliographystyle{ieee_fullname}
\bibliography{egbib}

\begin{thebibliography}{10}\itemsep=-1pt

\bibitem{DBLP:conf/icassp/AnwarHS15}
Sajid Anwar, Kyuyeon Hwang, and Wonyong Sung.
\newblock Fixed point optimization of deep convolutional neural networks for
  object recognition.
\newblock In {\em ICASSP}, pages 1131--1135, 2015.

\bibitem{DBLP:conf/icml/BenderKZVL18}
Gabriel Bender, Pieter{-}Jan Kindermans, Barret Zoph, Vijay Vasudevan, and
  Quoc~V. Le.
\newblock Understanding and simplifying one-shot architecture search.
\newblock In {\em ICML}, pages 549--558, 2018.

\bibitem{DBLP:conf/iclr/BrockLRW18}
Andrew Brock, Theodore Lim, James~M. Ritchie, and Nick Weston.
\newblock {SMASH:} one-shot model architecture search through hypernetworks.
\newblock In {\em ICLR}, 2018.

\bibitem{DBLP:conf/cvpr/CaiHSV17}
Zhaowei Cai, Xiaodong He, Jian Sun, and Nuno Vasconcelos.
\newblock Deep learning with low precision by half-wave gaussian quantization.
\newblock In {\em CVPR}, pages 5406--5414, 2017.

\bibitem{DBLP:conf/cvpr/CaiV18}
Zhaowei Cai and Nuno Vasconcelos.
\newblock Cascade {R-CNN:} delving into high quality object detection.
\newblock In {\em CVPR}, pages 6154--6162, 2018.

\bibitem{DBLP:journals/pami/ChenPKMY18}
Liang{-}Chieh Chen, George Papandreou, Iasonas Kokkinos, Kevin Murphy, and
  Alan~L. Yuille.
\newblock Deeplab: Semantic image segmentation with deep convolutional nets,
  atrous convolution, and fully connected crfs.
\newblock {\em {IEEE} Trans. Pattern Anal. Mach. Intell.}, 40(4):834--848,
  2018.

\bibitem{choi2018pact}
Jungwook Choi, Zhuo Wang, Swagath Venkataramani, Pierce I-Jen Chuang,
  Vijayalakshmi Srinivasan, and Kailash Gopalakrishnan.
\newblock Pact: Parameterized clipping activation for quantized neural
  networks.
\newblock {\em arXiv preprint arXiv:1805.06085}, 2018.

\bibitem{DBLP:journals/anor/ColsonMS07}
Beno{\^{\i}}t Colson, Patrice Marcotte, and Gilles Savard.
\newblock An overview of bilevel optimization.
\newblock {\em Annals {OR}}, 153(1):235--256, 2007.

\bibitem{DBLP:conf/iccv/DongYGMK19}
Zhen Dong, Zhewei Yao, Amir Gholami, Michael~W. Mahoney, and Kurt Keutzer.
\newblock {HAWQ:} hessian aware quantization of neural networks with
  mixed-precision.
\newblock In {\em ICCV}, pages 293--302. {IEEE}, 2019.

\bibitem{guo2019single}
Zichao Guo, Xiangyu Zhang, Haoyuan Mu, Wen Heng, Zechun Liu, Yichen Wei, and
  Jian Sun.
\newblock Single path one-shot neural architecture search with uniform
  sampling.
\newblock {\em arXiv preprint arXiv:1904.00420}, 2019.

\bibitem{DBLP:conf/iccv/HeGDG17}
Kaiming He, Georgia Gkioxari, Piotr Doll{\'{a}}r, and Ross~B. Girshick.
\newblock Mask {R-CNN}.
\newblock In {\em ICCV}, pages 2980--2988, 2017.

\bibitem{DBLP:journals/corr/HeZRS15}
Kaiming He, Xiangyu Zhang, Shaoqing Ren, and Jian Sun.
\newblock Deep residual learning for image recognition.
\newblock In {\em CVPR}, pages 770--778, 2016.

\bibitem{DBLP:journals/corr/HowardZCKWWAA17}
Andrew~G. Howard, Menglong Zhu, Bo Chen, Dmitry Kalenichenko, Weijun Wang,
  Tobias Weyand, Marco Andreetto, and Hartwig Adam.
\newblock Mobilenets: Efficient convolutional neural networks for mobile vision
  applications.
\newblock {\em CoRR}, abs/1704.04861, 2017.

\bibitem{DBLP:conf/nips/HubaraCSEB16}
Itay Hubara, Matthieu Courbariaux, Daniel Soudry, Ran El{-}Yaniv, and Yoshua
  Bengio.
\newblock Binarized neural networks.
\newblock In {\em NIPS}, pages 4107--4115, 2016.

\bibitem{DBLP:conf/sips/HwangS14}
Kyuyeon Hwang and Wonyong Sung.
\newblock Fixed-point feedforward deep neural network design using weights +1,
  0, and -1.
\newblock In {\em {IEEE} Workshop on Signal Processing Systems}, pages
  174--179, 2014.

\bibitem{DBLP:conf/nips/KrizhevskySH12}
Alex Krizhevsky, Ilya Sutskever, and Geoffrey~E. Hinton.
\newblock Imagenet classification with deep convolutional neural networks.
\newblock In {\em NIPS}, pages 1106--1114, 2012.

\bibitem{DBLP:conf/uai/LaceyTA18}
Griffin Lacey, Graham~W. Taylor, and Shawki Areibi.
\newblock Stochastic layer-wise precision in deep neural networks.
\newblock In {\em UAI}, pages 663--672, 2018.

\bibitem{DBLP:conf/icml/LinTA16}
Darryl~Dexu Lin, Sachin~S. Talathi, and V.~Sreekanth Annapureddy.
\newblock Fixed point quantization of deep convolutional networks.
\newblock In {\em ICML}, pages 2849--2858, 2016.

\bibitem{DBLP:conf/cvpr/LinDGHHB17}
Tsung{-}Yi Lin, Piotr Doll{\'{a}}r, Ross~B. Girshick, Kaiming He, Bharath
  Hariharan, and Serge~J. Belongie.
\newblock Feature pyramid networks for object detection.
\newblock In {\em CVPR}, pages 936--944, 2017.

\bibitem{liu2018darts}
Hanxiao Liu, Karen Simonyan, and Yiming Yang.
\newblock Darts: Differentiable architecture search.
\newblock {\em arXiv preprint arXiv:1806.09055}, 2018.

\bibitem{DBLP:journals/tit/Lloyd82}
Stuart~P. Lloyd.
\newblock Least squares quantization in {PCM}.
\newblock {\em {IEEE} Trans. Information Theory}, 28(2):129--136, 1982.

\bibitem{DBLP:journals/tit/Max60}
Joel Max.
\newblock Quantizing for minimum distortion.
\newblock {\em {IRE} Trans. Information Theory}, 6(1):7--12, 1960.

\bibitem{DBLP:conf/eccv/RastegariORF16}
Mohammad Rastegari, Vicente Ordonez, Joseph Redmon, and Ali Farhadi.
\newblock Xnor-net: Imagenet classification using binary convolutional neural
  networks.
\newblock In {\em ECCV}, pages 525--542, 2016.

\bibitem{DBLP:conf/nips/RenHGS15}
Shaoqing Ren, Kaiming He, Ross~B. Girshick, and Jian Sun.
\newblock Faster {R-CNN:} towards real-time object detection with region
  proposal networks.
\newblock In {\em NIPS}, pages 91--99, 2015.

\bibitem{DBLP:journals/ijcv/RussakovskyDSKS15}
Olga Russakovsky, Jia Deng, Hao Su, Jonathan Krause, Sanjeev Satheesh, Sean Ma,
  Zhiheng Huang, Andrej Karpathy, Aditya Khosla, Michael~S. Bernstein,
  Alexander~C. Berg, and Fei{-}Fei Li.
\newblock Imagenet large scale visual recognition challenge.
\newblock {\em International Journal of Computer Vision}, 115(3):211--252,
  2015.

\bibitem{DBLP:journals/corr/SimonyanZ14a}
Karen Simonyan and Andrew Zisserman.
\newblock Very deep convolutional networks for large-scale image recognition.
\newblock {\em CoRR}, abs/1409.1556, 2014.

\bibitem{sun2008image}
Huifang Sun and Yun~Q Shi.
\newblock {\em Image and video compression for multimedia engineering:
  Fundamentals, algorithms, and standards}.
\newblock CRC press, 2008.

\bibitem{DBLP:journals/tsp/SungK95}
Wonyong Sung and Ki{-}Il Kum.
\newblock Simulation-based word-length optimization method for fixed-point
  digital signal processing systems.
\newblock {\em {IEEE} Trans. Signal Processing}, 43(12):3087--3090, 1995.

\bibitem{DBLP:conf/cvpr/SzegedyLJSRAEVR15}
Christian Szegedy, Wei Liu, Yangqing Jia, Pierre Sermanet, Scott~E. Reed,
  Dragomir Anguelov, Dumitru Erhan, Vincent Vanhoucke, and Andrew Rabinovich.
\newblock Going deeper with convolutions.
\newblock In {\em CVPR}, pages 1--9, 2015.

\bibitem{DBLP:conf/cvpr/SzegedyVISW16}
Christian Szegedy, Vincent Vanhoucke, Sergey Ioffe, Jonathon Shlens, and
  Zbigniew Wojna.
\newblock Rethinking the inception architecture for computer vision.
\newblock In {\em CVPR}, pages 2818--2826, 2016.

\bibitem{wang2019haq}
Kuan Wang, Zhijian Liu, Yujun Lin, Ji Lin, and Song Han.
\newblock Haq: Hardware-aware automated quantization with mixed precision.
\newblock In {\em CVPR}, pages 8612--8620, 2019.

\bibitem{wu2018mixed}
Bichen Wu, Yanghan Wang, Peizhao Zhang, Yuandong Tian, Peter Vajda, and Kurt
  Keutzer.
\newblock Mixed precision quantization of convnets via differentiable neural
  architecture search.
\newblock {\em arXiv preprint arXiv:1812.00090}, 2018.

\bibitem{yazdanbakhsh2018releq}
Amir Yazdanbakhsh, Ahmed~T Elthakeb, Prannoy Pilligundla, FatemehSadat
  Mireshghallah, and Hadi Esmaeilzadeh.
\newblock Releq: An automatic reinforcement learning approach for deep
  quantization of neural networks.
\newblock {\em arXiv preprint arXiv:1811.01704}, 2018.

\bibitem{DBLP:conf/eccv/ZhangYYH18}
Dongqing Zhang, Jiaolong Yang, Dongqiangzi Ye, and Gang Hua.
\newblock Lq-nets: Learned quantization for highly accurate and compact deep
  neural networks.
\newblock In {\em ECCV}, pages 373--390, 2018.

\bibitem{DBLP:journals/corr/ZhouNZWWZ16}
Shuchang Zhou, Zekun Ni, Xinyu Zhou, He Wen, Yuxin Wu, and Yuheng Zou.
\newblock Dorefa-net: Training low bitwidth convolutional neural networks with
  low bitwidth gradients.
\newblock {\em CoRR}, abs/1606.06160, 2016.

\bibitem{zhuang2019structured}
Bohan Zhuang, Chunhua Shen, Mingkui Tan, Lingqiao Liu, and Ian Reid.
\newblock Structured binary neural networks for image recognition.
\newblock {\em arXiv preprint arXiv:1909.09934}, 2019.

\bibitem{DBLP:conf/cvpr/ZhuangSTL018}
Bohan Zhuang, Chunhua Shen, Mingkui Tan, Lingqiao Liu, and Ian~D. Reid.
\newblock Towards effective low-bitwidth convolutional neural networks.
\newblock In {\em CVPR}, pages 7920--7928, 2018.

\bibitem{DBLP:journals/corr/ZophL16}
Barret Zoph and Quoc~V. Le.
\newblock Neural architecture search with reinforcement learning.
\newblock {\em CoRR}, abs/1611.01578, 2016.

\bibitem{DBLP:conf/cvpr/ZophVSL18}
Barret Zoph, Vijay Vasudevan, Jonathon Shlens, and Quoc~V. Le.
\newblock Learning transferable architectures for scalable image recognition.
\newblock In {\em CVPR}, pages 8697--8710, 2018.

\end{thebibliography}
}

\end{document}